\documentclass{article}

\usepackage[a4paper, top=2.5cm, bottom=2.5cm, left=2cm, right=2cm]{geometry}
\usepackage[utf8]{inputenc}
\usepackage[T1]{fontenc}
\usepackage{lmodern}
\usepackage[english]{babel}
\usepackage{amsmath, amssymb, amsthm, mathtools}
\usepackage{graphicx}
\usepackage{booktabs}
\usepackage{hyperref}
\usepackage{subcaption}
\usepackage{cite}
\usepackage{algorithm}
\usepackage{algpseudocode}
\usepackage{xcolor}
\usepackage{float}

\hypersetup{
    colorlinks=true,
    linkcolor=blue,
    filecolor=magenta,      
    urlcolor=cyan,
    citecolor=red,
}

\newcommand{\Loss}{\mathcal{L}}

\title{\textbf{Grokking as a Variance-Limited Phase Transition:}\\ Spectral Gating and the Epsilon-Stability Threshold}

\author{
  \textbf{Pratyush Acharya} \\
  \texttt{acharya.pratyush12@gmail.com}
    \and
  \textbf{Habish Dhakal} \\
  \texttt{dhakalhabish@gmail.com}
}
\date{\today}

\begin{document}

\maketitle

\begin{abstract}
Standard optimization theories struggle to explain grokking, where generalization occurs long after training convergence. While geometric studies attribute this to slow drift, they often overlook the interaction between the optimizer's noise structure and landscape curvature. This work analyzes AdamW dynamics on modular arithmetic tasks, revealing a ``Spectral Gating'' mechanism that regulates the transition from memorization to generalization. 

We find that AdamW operates as a variance-gated stochastic system. Grokking is constrained by a stability condition: the generalizing solution resides in a sharp basin ($\lambda_{max}^H$) initially inaccessible under low-variance regimes. The ``delayed'' phase represents the accumulation of gradient variance required to lift the effective stability ceiling, permitting entry into this sharp manifold. 

Our ablation studies identify three complexity regimes: (1) \textbf{Capacity Collapse} ($P < 23$), where rank-deficiency prevents structural learning; (2) \textbf{The Variance-Limited Regime} ($P \approx 41$), where generalization waits for the spectral gate to open; and (3) \textbf{Stability Override} ($P > 67$), where memorization becomes dimensionally unstable. Furthermore, we challenge the "Flat Minima" hypothesis for algorithmic tasks, showing that isotropic noise injection fails to induce grokking. Generalization requires the \textit{anisotropic rectification} unique to adaptive optimizers, which directs noise into the tangent space of the solution manifold.
\end{abstract}

\section{Introduction}

Sometimes, a neural network learns a task only after it has already perfectly fit the training data. This delayed phase transition, known as \textit{grokking} \cite{power2022grokking}, challenges the traditional view that optimization ends when gradients hit zero. Understanding this disconnect is vital, as it may explain why large foundation models continue to improve reasoning capabilities even after their training loss plateaus. This implies the existence of a high-dimensional \textbf{Minimizing Level Set} \cite{musat2025geometry, pesme2025theoretical} $\mathcal{Z} = \{ \theta \mid \Loss(\theta) \approx 0 \}$, where dynamics are driven not by the loss gradient, but by the optimizer's internal noise structure.

Earlier studies explain these dynamics in three ways:
\begin{enumerate}
    \item \textbf{Geometric Drift:} Theories suggesting that weight decay slowly pushes the model toward max-margin solutions \cite{musat2025geometry, pesme2025theoretical}. While mathematically rigorous, these frameworks assume a generic gradient flow. They often fail to explain why isotropic noise (SGLD) or standard SGD does not grok, even when weight decay is present \cite{thilak2022slingshot}.
    
    \item \textbf{Circuit Competition:} Mechanistic interpretations that view grokking as a battle between ``memorizing'' and ``generalizing'' circuits \cite{nanda2023progress, varma2023explaining}. These studies map the \textit{topology} of the solution (e.g., the ``Clock Circuit'') but lack a kinetic mechanism to explain the timescale of the transition.
    
    \item \textbf{Edge of Stability (EoS):} Research arguing that instability drives feature learning \cite{cohen2021gradient, kumar2024grokking}. However, the precise interaction between numerical stability parameters (such as $\epsilon$) and the grokking phase transition remains under-explored.
\end{enumerate}

In this paper, we unify these perspectives by analyzing the optimizer as a \textbf{Variance-Gated Stochastic System}. We find that grokking is a fragile state that depends on matching task complexity to optimizer stability. We posit that the delay in generalization is not a random walk, but a \textbf{Variance-Limited Equilibrium}. The optimizer is initially too stable to enter the sharp manifolds where generalizing circuits reside.

Our contributions are as follows:
\begin{itemize}
    \item \textbf{The Spectral Gating Mechanism:} We find that the generalizing basin for modular arithmetic tasks is significantly sharper than the optimizer's initial stability threshold. Grokking appears to occur only when accumulated gradient variance $\sqrt{v_t}$ lifts the stability ceiling ($2/\eta_{eff}$), finally permitting entry into the solution's high-curvature geometry.
    
    \item \textbf{The Complexity Threshold:} We identify a ``Signal Starvation'' regime ($P < 23$) where the loss landscape is structurally barren. Below this threshold, neither variance accumulation nor parameter tuning induces generalization, refuting purely thermal explanations of grokking.
    
    \item \textbf{Anisotropic Rectification:} By comparing AdamW to Isotropic SGLD, we show that thermal energy alone is insufficient. Generalization requires the covariance structure of adaptive optimizers to rectify noise into the tangent space of the solution manifold.
\end{itemize}

\section{Related Work}

\paragraph{Grokking and Phase Transitions.}
The phenomenon of grokking was first characterized by Power et al. \cite{power2022grokking} as generalization occurring long after training error converges to zero. Subsequent work has attempted to demarcate the conditions for this transition. Liu et al. \cite{liu2022omnigrok} identified a "Goldilocks zone" of initialization and data size, while Kumar et al. \cite{kumar2024grokking} framed it as a transition from "lazy" kernel regimes to "rich" feature learning.
While these works describe the \textit{phenomenology} of the transition, they largely treat the optimizer as a black box. Our work differs by providing a \textit{kinetic} explanation: we identify the specific spectral state the optimizer must reach to trigger the transition.

\paragraph{Mechanistic Interpretability.}
A parallel line of inquiry focuses on \textit{what} is learned during grokking. Nanda et al. \cite{nanda2023progress} fully reverse-engineered the modular addition network, identifying a specific "Clock Circuit" relying on trigonometric interference. Varma et al. \cite{varma2023explaining} and Merrill et al. \cite{merrill2023tale} argue that grokking is a competition between dense memorization circuits and sparse generalizing circuits.
We adopt this topological view—that the solution is a specific, sparse circuit—but address the open question of \textit{accessibility}. While mechanistic work maps the destination, our spectral gating theory explains the optimization dynamics required to enter the destination basin.

\paragraph{Implicit Bias and the Edge of Stability.}
Geometric theories posit that gradient descent introduces an implicit bias toward max-margin solutions \cite{soudry2018implicit}. Recent rigorous frameworks by Musat \cite{musat2025geometry} and Pesme et al. \cite{pesme2025theoretical} model grokking as Riemannian Norm Minimization on the zero-loss manifold.
However, these "drift" theories often assume stable gradient flow. Cohen et al. \cite{cohen2021gradient} and Damian et al. \cite{damian2023self} demonstrate that neural network training typically occurs at the "Edge of Stability," where the sharpness oscillates around $2/\eta$. Thilak et al. \cite{thilak2022slingshot} empirically linked this instability ("Slingshots") to grokking.
Our work unifies the geometric and stability perspectives. We formalize the Slingshot not as an anomaly, but as a \textit{variance-injection mechanism} required to satisfy the stability constraints of the sharp basins identified by mechanistic interpretability. Furthermore, we challenge the prevailing "Flat Minima" hypothesis \cite{hochreiter1997flat, xie2021diffusion} in this specific domain, providing evidence that for algorithmic tasks, the generalizing solution is spectrally sharper than the memorization solution.

\section{Theoretical Framework: Spectral Gating and Rectified Stochastic Dynamics}

To understand how AdamW moves through the loss landscape, we model its training dynamics as a continuous stochastic process. While classical optimization focuses on the convergence of the loss mean, grokking takes place in a "post-convergence" regime where $\mathcal{L}(\theta) \to 0$ yet the weights $\|\theta\|$ continue to evolve \cite{power2022grokking}. In this phase, the trajectory is governed not by the loss gradient, which is near zero, but by the \textbf{noise covariance structure} of the optimizer.

\subsection{AdamW as Variance-Gated Stochastic Dynamics}

We treat the discrete updates of AdamW as a continuous-time Stochastic Differential Equation (SDE) to capture the interaction between noise and geometry. Let $g_t(\theta) = \nabla \mathcal{L}(\theta) + \xi_t$ be the stochastic gradient, where $\xi_t \sim \mathcal{N}(0, \Sigma(\theta_t))$ represents anisotropic, often heavy-tailed noise \cite{simsekli2019tail, xie2021diffusion}. 

AdamW preconditions these updates using $v_t$, the exponential moving average of squared gradients. Under the \textbf{Adiabatic Approximation} \cite{kunstner2019limitations}, we assume the preconditioner $D_t = \text{diag}(\sqrt{v_t} + \epsilon)$ evolves on a slower timescale than the parameters. Matching the first and second moments of the discrete step $\Delta \theta \approx -\eta D_t^{-1} g_t$ yields the following SDE \cite{balles2018dissecting}:

\begin{equation}
\mathrm{d}\theta_t = \underbrace{- D_t^{-1} \left( \nabla \mathcal{L}(\theta_t) + \lambda \theta_t \right) \mathrm{d}t}_{\text{Preconditioned Drift}} + \underbrace{\eta^{1/2} D_t^{-1} \Sigma(\theta_t)^{1/2} \mathrm{d}W_t}_{\text{Rectified Diffusion}}
\end{equation}

where $W_t$ is a standard Brownian motion. The $\eta^{1/2}$ scaling ensures the diffusion term matches the $\mathcal{O}(\eta)$ variance of the discrete algorithm.

Crucially, AdamW behaves differently from standard Riemannian Langevin Dynamics (RLD). While RLD scales diffusion by the inverse square root of the metric ($D_t^{-1/2}$), AdamW scales it by the inverse ($D_t^{-1}$). This structural difference creates a distinct diffusivity profile for the $i$-th parameter:

\begin{equation}
\mathcal{D}_{eff}^{(i)} \propto \frac{\eta \cdot \Sigma_{ii}}{(\sqrt{\Sigma_{ii}} + \epsilon)^2}
\end{equation}

As the gradient noise variance $\Sigma_{ii} \to \infty$, RLD diffusion becomes unbounded. In contrast, AdamW diffusion saturates at a fixed limit ($\mathcal{D}_{eff} \to \eta$). This \textbf{Bounded Diffusivity} ensures that high-variance gradients do not cause the system to diverge from the manifold, effectively creating a "variance ceiling" that maintains stability even in high-noise regimes.

\textit{Remark on Discrete Dynamics:} While the SDE provides an equilibrium description, the transition into the grokking phase is often triggered by "Slingshots" \cite{thilak2022slingshot}—discrete instabilities where the adiabatic assumption momentarily breaks down. In these transient moments, the rapid accumulation of $v_t$ (as described in Section \ref{sec:spectral_gating}) serves to actively suppress the effective step size, dynamically restoring stability.

\subsection{The Stability-Diffusion Trade-off}

Equation (2) shows how the stability parameter $\epsilon$ regulates the effective diffusion coefficient. By analyzing the limits of this equation, we can derive the three distinct physical regimes observed in our experiments.

\subsubsection{Regime I: Variance Suppression (Over-Damped)}
\textbf{Condition:} $\epsilon \gg \sigma_i$.
When the stability term dominates the preconditioner, the effective diffusivity vanishes:
\begin{equation}
\lim_{\epsilon \to \infty} \mathcal{D}_{eff}^{(i)} \approx \frac{\eta \sigma_i^2}{\epsilon^2} \to 0
\end{equation}
In this limit, AdamW degenerates into \textbf{Damped SGD}. The diffusive pressure becomes too weak to overcome the restorative drift of weight decay \cite{loshchilov2017decoupled}. As seen in the phase diagram (Figure \ref{fig:heatmap}), this results in model stagnation; the system remains trapped in the kernel regime, unable to explore the manifold.

\subsubsection{Regime II: Radial Expansion (Under-Damped)}
\textbf{Condition:} $\epsilon \ll \sigma_i$.
When $\epsilon$ is negligible, the preconditioner perfectly cancels the noise magnitude:
\begin{equation}
\lim_{\epsilon \to 0} \mathcal{D}_{eff}^{(i)} \approx \frac{\eta \sigma_i^2}{(\sqrt{\sigma_i^2})^2} = \eta
\end{equation}
While this maximizes exploration, it removes the numerical floor required for stability. In our ablation studies, setting $\epsilon \to 10^{-15}$ caused the weight norm $\|\theta\|^2$ to diverge. Without the $\epsilon$ constraint, the system enters a phase of \textbf{Radial Expansion}, diffusing outwardly faster than weight decay can pull it back \cite{pesme2025theoretical}.

\subsubsection{Regime III: Anisotropic Rectification (Grokking)}
\textbf{Condition:} $\epsilon \approx \sigma_{noise}$.
Grokking occurs when $\epsilon$ is balanced. In this band, the optimizer performs \textbf{Anisotropic Rectification}: it amplifies noise in directions where the signal is coherent (low $\sigma$) while damping chaotic directions (high $\sigma$). This balance allows the system to drift along the tangent space of $\mathcal{Z}$ without diverging.

\subsection{Spectral Gating and the Edge of Stability}

For intermediate complexity tasks ($P \approx 23-60$), we observe that structural learning is delayed by a \textbf{Spectral Lock}. We explain this by connecting our framework to the \textbf{Edge of Stability (EoS)} theory \cite{cohen2021gradient}.

In discrete optimization, a step size $\eta$ is stable only if the local curvature satisfies $\lambda < 2/\eta$. For AdamW, the effective step size is adaptive and parameter-specific: $\eta_{eff} \approx \eta / (\sqrt{v_t} + \epsilon)$. Consequently, the stability condition becomes dynamic. The optimizer can only converge into a basin with Hessian curvature $\lambda_{max}^H$ if it satisfies the \textbf{Spectral Gating Condition}:

\begin{equation}
\lambda_{max}^H < \frac{2}{\eta} (\sqrt{v_t} + \epsilon)
\end{equation}

This inequality uncovers the causal mechanism behind the delay:

\begin{enumerate}
    \item \textbf{The Memorization Trap (Low Variance):} Initially, gradient variance $v_t$ is low. This suppresses the stability ceiling ($2/\eta_{eff}$), forcing the optimizer to seek flat, high-entropy minima. The "lazy" memorization solution fits this profile: it is broad and robust to small perturbations, making it immediately accessible.
    
    \item \textbf{The Sharpness Inversion:} Contrary to the "Flat Minima" hypothesis \cite{hochreiter1997flat}, we argue that for algorithmic tasks, the generalizing solution (the "Clock Circuit" \cite{nanda2023progress}) is geometrically \textit{sharper} than the memorization basin. It requires precise parameter alignment, resulting in a high $\lambda_{max}^H$ that initially violates the condition in Eq. (5).
    
    \item \textbf{Stability Release (Variance Injection):} As training progresses, gradients do not vanish but fluctuate, accumulating variance $v_t$ (the "Slingshot" mechanism). This increase in the denominator of the update rule effectively anneals the step size. This lifts the stability ceiling defined by Eq. (5), finally permitting the optimizer to stably enter and settle in the sharp generalizing manifold.
\end{enumerate}

Thus, the delay is not a random walk, but a \textbf{variance-accumulation phase}. The model must generate enough gradient noise to "unlock" the spectral gate, transitioning from the variance-intolerant memorization basin to the variance-stabilized generalizing circuit.

\section{Experimental Setup}

\textbf{Task \& Model.} Our experiments utilize an MLP with 2 hidden layers (width 128, ReLU activations) trained on the modular addition task $a + b \pmod P$ with embeddings learned from scratch. Training employs full-batch AdamW with learning rate $\eta = 10^{-4}$, weight decay $\lambda = 0.1$, and $\beta=(0.9, 0.999)$ unless specified otherwise.

\textbf{Protocol.} We perform three targeted experimental sweeps to dissect the underlying mechanism:

\begin{enumerate}
    \item \textbf{The Stability Frontier (Hardness vs. Gating):} To map the thermodynamic boundaries of grokking, we conduct a dense 2D grid search over Task Difficulty ($P \in [11, 97]$, sampled linearly) and Optimizer Stability ($\epsilon \in [10^{-9}, 10^{-1}]$, sampled logarithmically). Each configuration is trained for $10^6$ steps to determine the "Steps to Grok" (defined as Test Accuracy $> 99\%$).
    
    \item \textbf{Thermodynamic Sufficiency (SGLD):} To determine if isotropic energy can substitute for geometric steering, we compare AdamW against Stochastic Gradient Langevin Dynamics (SGLD). By injecting Gaussian noise $\mathcal{N}(0, \sigma^2)$ into standard SGD updates ($\sigma \in [10^{-4}, 10^{-1}]$), we match the effective thermal pressure of AdamW for both easy ($P = 23$) and hard ($P = 67$) tasks.
    
    \item \textbf{Mechanism \& Topology:} Finally, we monitor high-resolution topological metrics, including the \textbf{Force Ratio} ($R = C_{push}/C_{pull}$), the \textbf{Hessian Trace} (via Hutchinson's estimator), and \textbf{Fourier Sparsity} (the $L_1$ norm of the weights' DFT). These metrics allow us to track the formation of the "Clock Circuit" structure in real-time.
\end{enumerate}

These experiments define the three dynamical regimes analyzed in the following section.

\section{Results and Analysis}

\subsection{Phase Boundaries of Generalization: The Impact of Task Complexity}
\label{sec:phase_diagram}

\begin{figure}[H]
    \centering
    \includegraphics[width=0.95\linewidth]{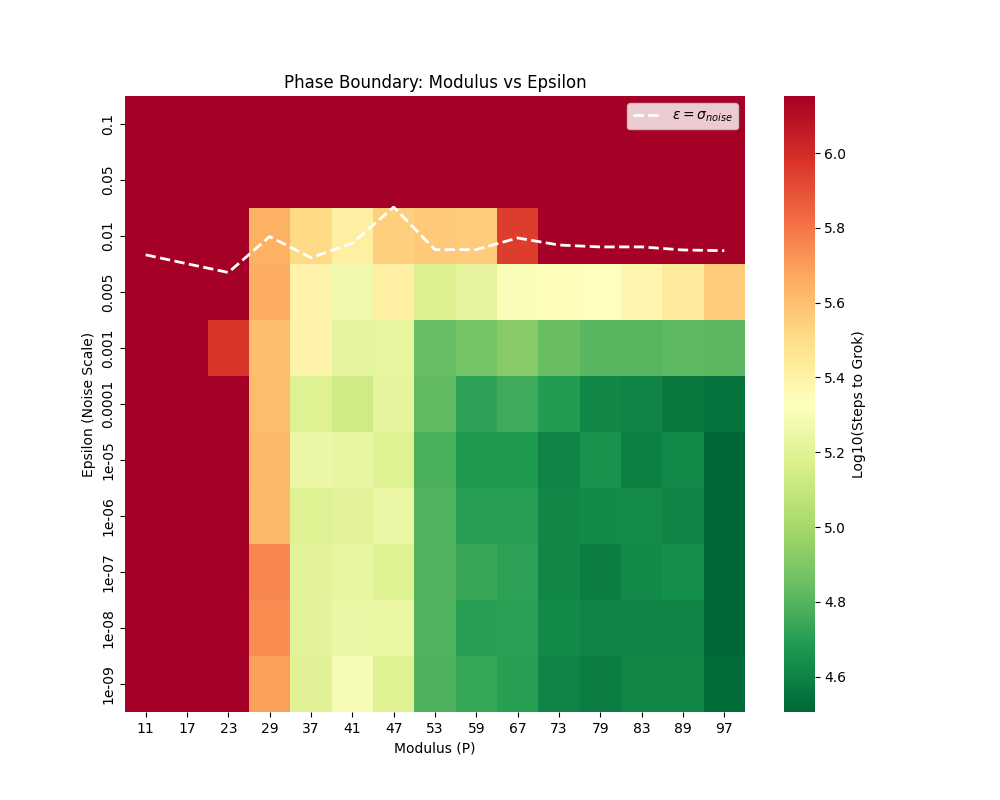}
    \caption{\textbf{Stability-Complexity Phase Diagram.} We define three regimes based on the time-to-generalization: (1) \textbf{Capacity Collapse} ($P < 23$), where rank-deficiency prevents the representation of the solution; (2) \textbf{Variance-Limited Regime} ($29 \le P \le 59$), where generalization is delayed by spectral gating; and (3) \textbf{Stability Override} ($P \ge 67$), where the high dimensionality of the memorization manifold forces immediate structural learning. The white dashed line marks $\epsilon = \sigma_{noise}$, representing the intrinsic gradient noise level.}
   \label{fig:heatmap}
\end{figure}

To map the dynamical regimes of grokking, we performed a dense grid search over Task Difficulty ($P \in [11, 97]$) and Optimizer Stability ($\epsilon \in [10^{-9}, 10^{-1}]$). The resulting phase diagram (Figure \ref{fig:heatmap}) reveals that generalization speed is non-monotonic with respect to complexity. Our results reveal three distinct regimes that explain how grokking begins and ends.

\subsubsection{Regime I: Capacity Collapse ($P < 23$)}
\begin{figure}[H]
    \centering
    \includegraphics[width=0.95\linewidth]{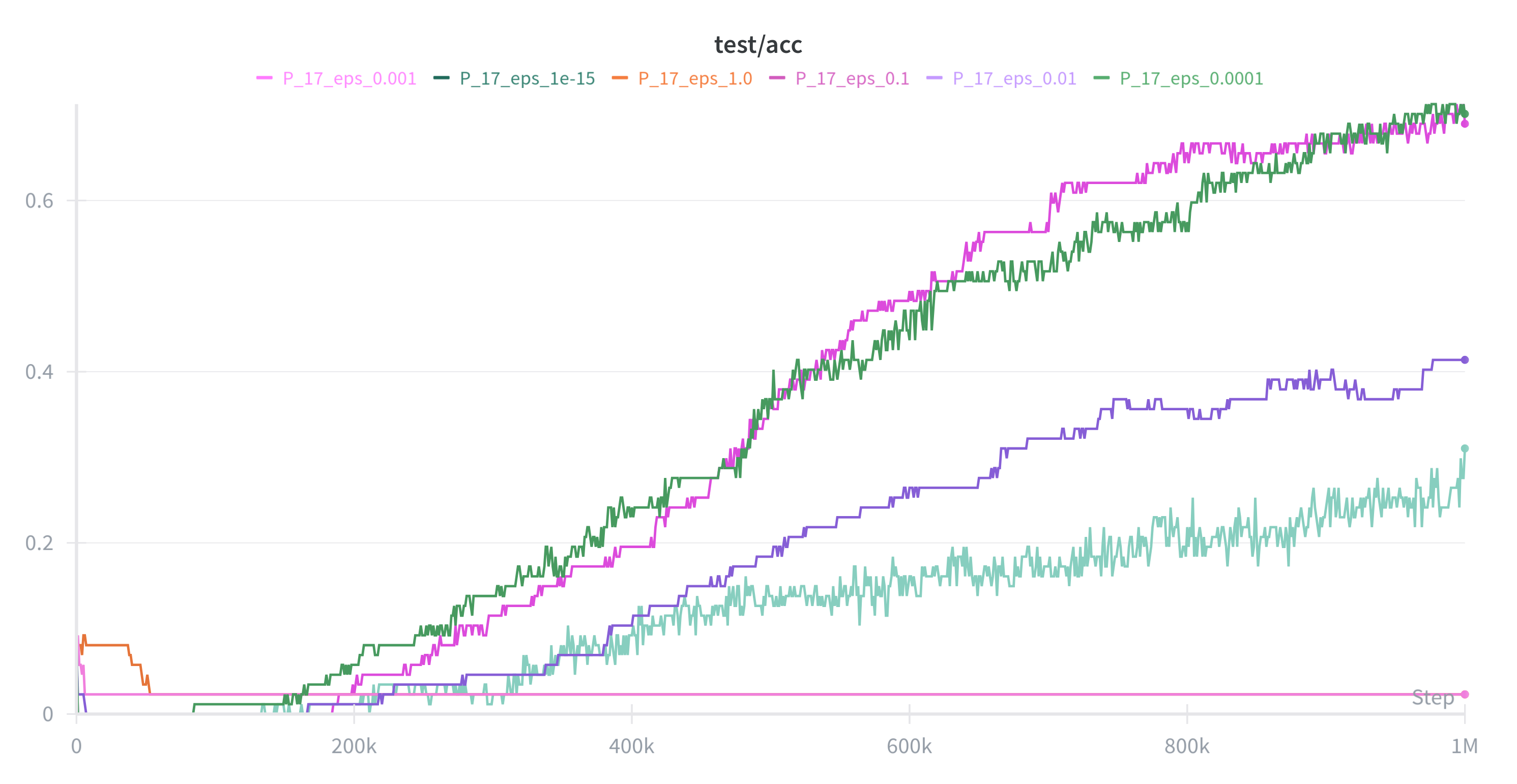}
    \caption{\textbf{Capacity Collapse ($P=17$).} Despite aggressive stability tuning ($\epsilon \to 10^{-15}$), the optimizer fails to locate the generalizing minimum. The dynamics exhibit stochastic tunneling without convergence, indicating that the embedding space lacks the geometric capacity to separate the task's Fourier modes.}
    \label{fig:p17_tunneling}
\end{figure}

The region $P < 23$ is characterized by convergence failure (Figure \ref{fig:heatmap}, Red). While often counter-intuitive that easier tasks are harder to learn, we attribute this to **Capacity Collapse** (or Rank Deficiency). As detailed in Section \ref{sec:capacity_collapse}, the modular addition task requires the model to represent $P$ distinct Fourier modes \cite{liu2022omnigrok}. When the embedding dimension $d_{model}$ is small relative to $P$ (or when $P$ is small enough that the orthogonality of high-frequency modes is compromised by initialization variance), the gradient signal becomes rank-deficient. The model enters a "tunneling" phase (Figure \ref{fig:p17_tunneling}) where it drifts stochastically without locking into a coherent minimum. This is a representational failure, not a thermodynamic one.

\subsubsection{Regime II: The Variance-Limited Regime ($29 \le P \le 59$)}
\begin{figure}[H]
    \centering
    \includegraphics[width=0.95\linewidth]{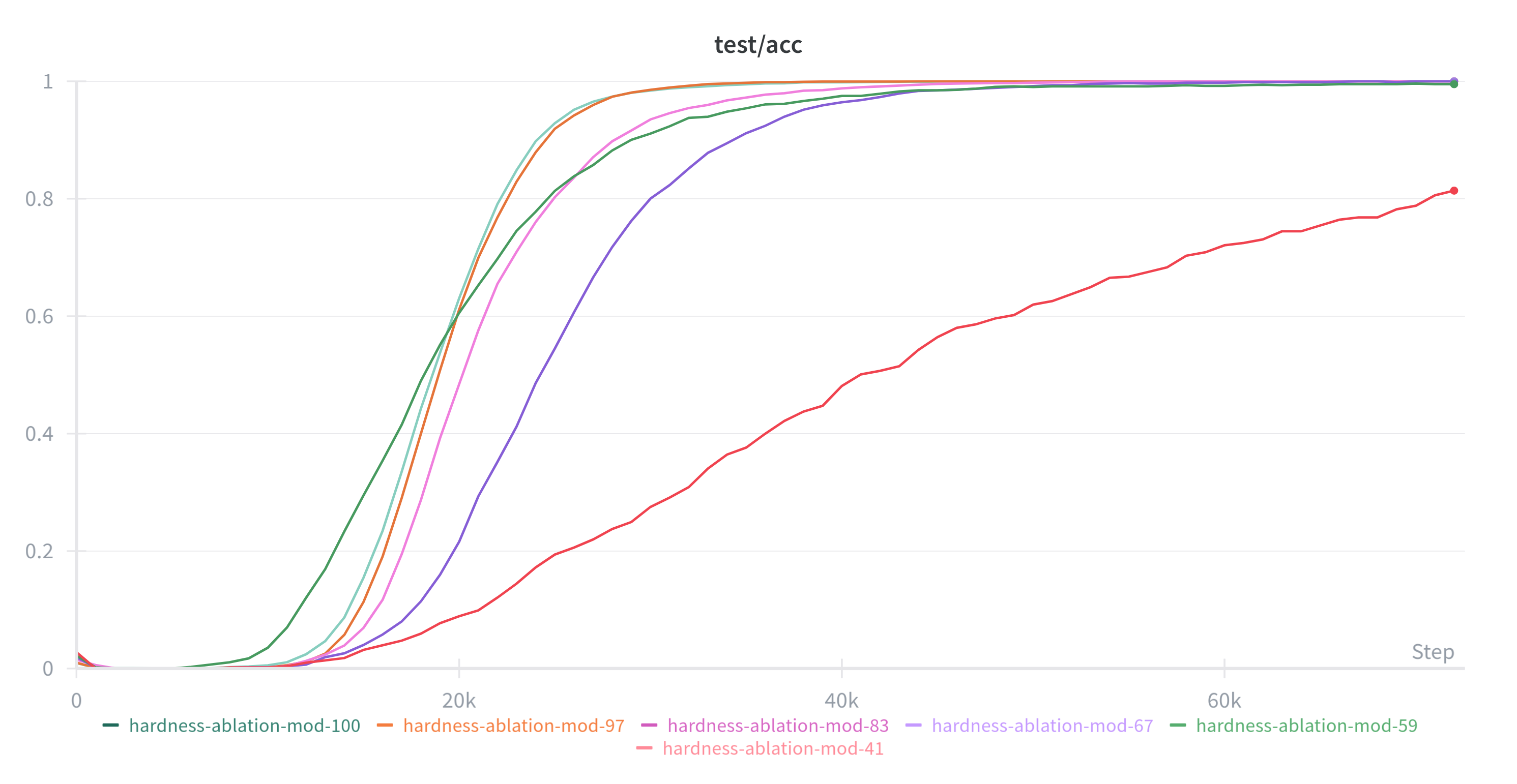} 
    \caption{\textbf{Non-Monotonic Generalization Dynamics.} Generalization time peaks at intermediate complexity ($P=41$, Red). Hard tasks ($P=97$, Teal) generalize immediately (Stability Override), while easy tasks ($P=23$, Purple) stagnate. The results reveal a "Complexity Valley" where the competition between the entropic pull of memorization and the spectral gate of generalization is maximized.}
    \label{fig:hardness_dynamics}
\end{figure}

Intermediate tasks exhibit the canonical "Grokking Gap," peaking at $P=41$ (Figure \ref{fig:hardness_dynamics}, Red line). We attribute this to **Competitive Variance Accumulation**. The memorization solution acts as a "Lazy" attractor: it is geometrically broad (high entropy) and easily accessible from random initialization. However, it is not the global minimum. The optimizer rapidly converges to this basin but is eventually destabilized by the accumulation of gradient variance $\sqrt{v_t}$. The delay corresponds to the time required for $v_t$ to grow sufficiently to "heat" the optimizer out of the flat memorization trap and into the sharper generalizing basin.

\subsubsection{Regime III: Stability Override ($P \ge 67$)}
For high-complexity tasks ($P \ge 67$), the grokking delay vanishes (Figure \ref{fig:hardness_dynamics}, Teal line). We observe **Stability Override**. The parameter cost of a memorization look-up table scales quadratically as $\mathcal{O}(P^2)$, whereas the generalizing circuit (a constant frequency rotation) remains $\mathcal{O}(1)$ \cite{varma2023explaining}. As $P$ increases, the volume of the parameter space occupied by valid memorization solutions shrinks exponentially relative to the structural solution. At $P=100$, the memorization basin becomes dimensionally unstable—it is simply too "small" to catch the optimizer. Consequently, AdamW bypasses the "Lazy" phase entirely, converging directly to the structural solution.

\subsection{Radial Stationarity: The Push-Pull Dynamics}

Post-convergence dynamics are governed by a radial equilibrium between Weight Decay ($\ell_2$ regularization) and the Rectified Diffusion of the optimizer. We model the weight norm $\|\theta\|^2$ as a radial Ornstein-Uhlenbeck process.

\begin{figure}[H]
    \centering
    \begin{subfigure}{0.32\textwidth}
        \centering
        \includegraphics[width=\linewidth]{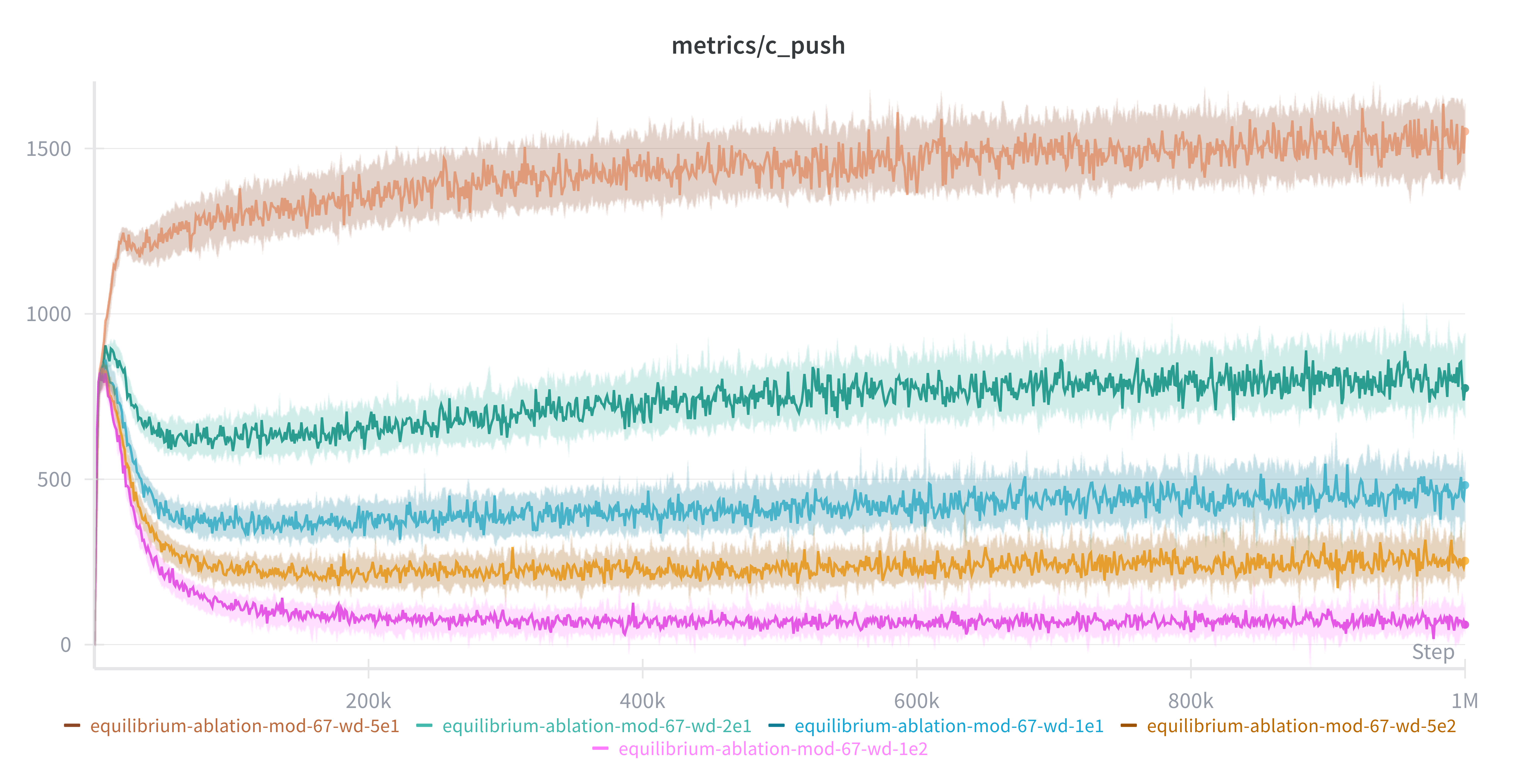}
        \caption{Diffusive Expansion ($F_{push}$)}
    \end{subfigure}
    \hfill
    \begin{subfigure}{0.32\textwidth}
        \centering
        \includegraphics[width=\linewidth]{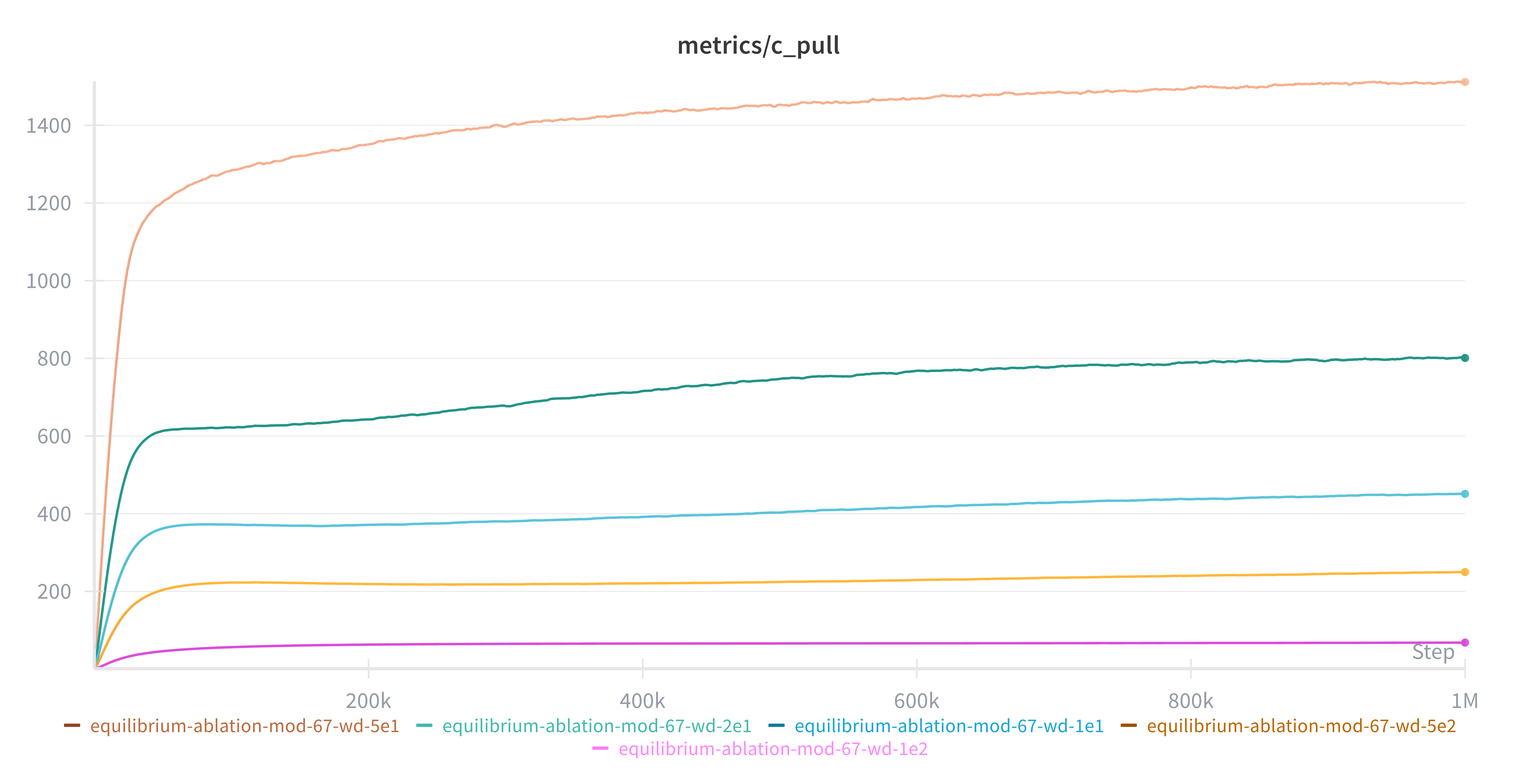}
        \caption{Restorative Drift ($F_{pull}$)}
    \end{subfigure}
    \hfill
    \begin{subfigure}{0.32\textwidth}
        \centering
        \includegraphics[width=\linewidth]{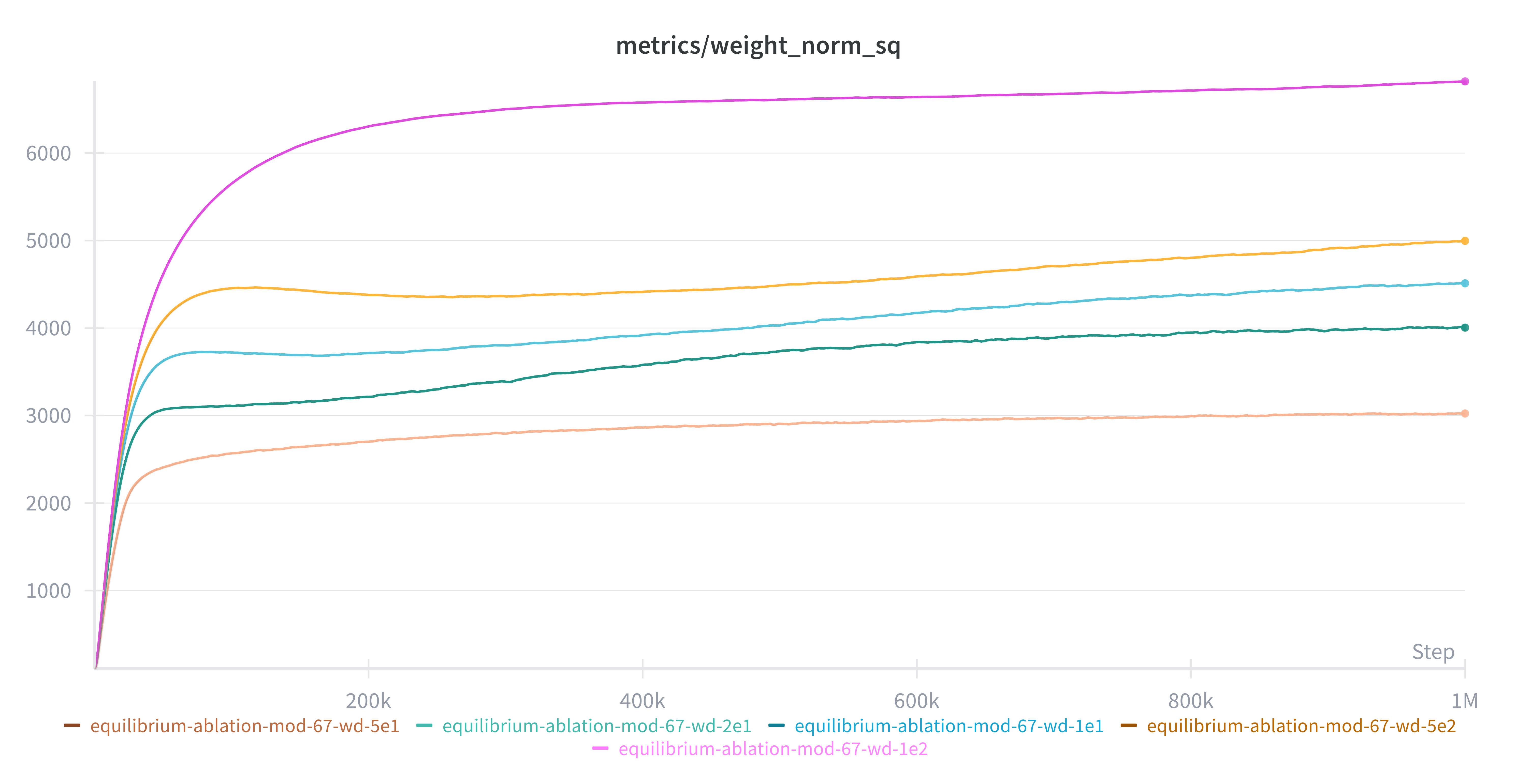}
        \caption{Radial Orbit $\|\theta\|^2$}
    \end{subfigure}
    \caption{\textbf{Radial Stationarity.} (a) AdamW generates a constant diffusive "push" (peach line) that counters the restorative force. (b) The restorative drift stabilizes, confirming a fixed radial orbit. (c) Stronger weight decay (peach) forces a lower-norm equilibrium, compressing the search space onto the manifold \cite{musat2025geometry}.}
    \label{fig:equilibrium_forces}
\end{figure}

\begin{figure}[H]
    \centering
    \begin{subfigure}{0.48\textwidth}
        \centering
        \includegraphics[width=\linewidth]{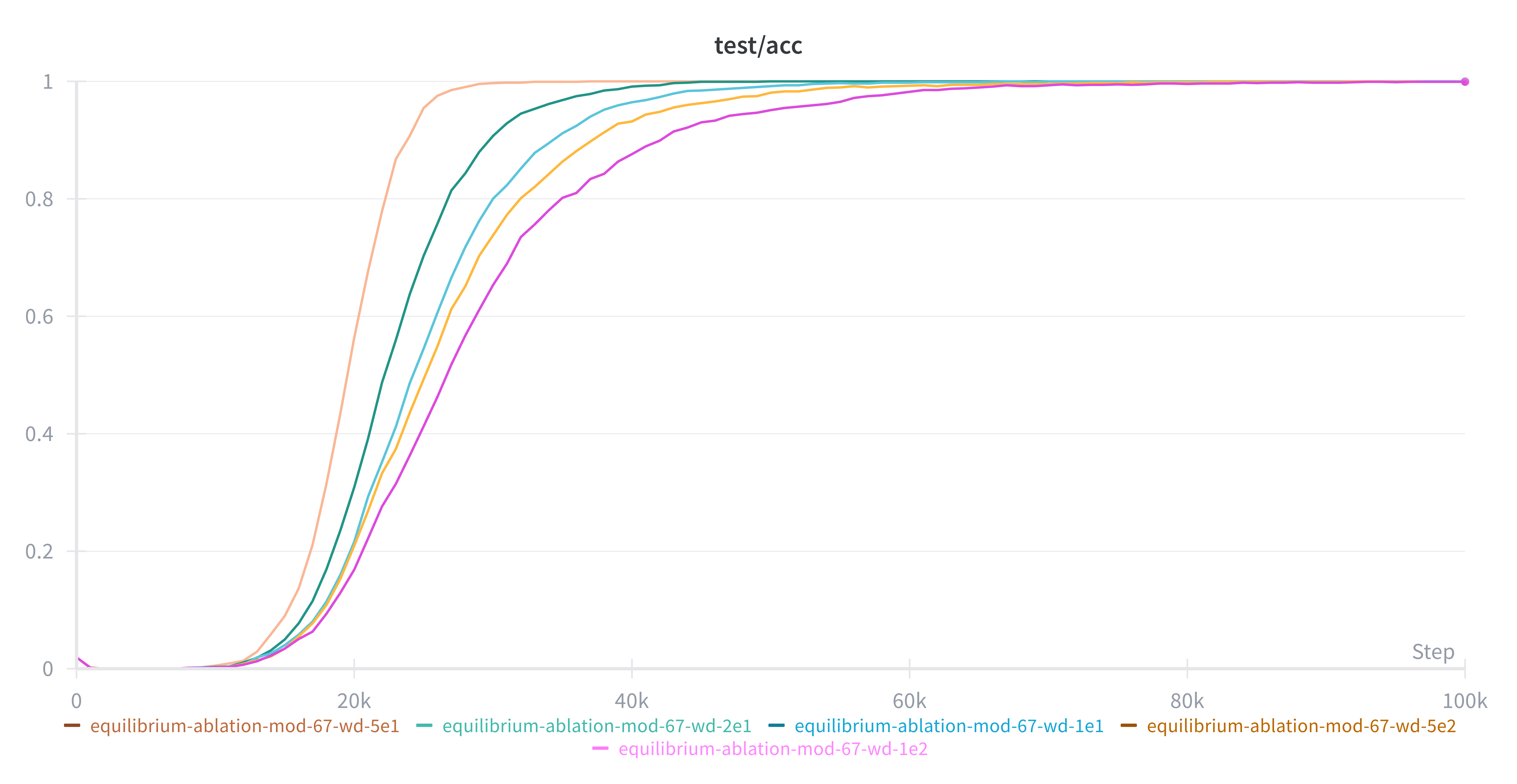}
        \caption{Test Accuracy vs. Steps}
    \end{subfigure}
    \hfill
    \begin{subfigure}{0.48\textwidth}
        \centering
        \includegraphics[width=\linewidth]{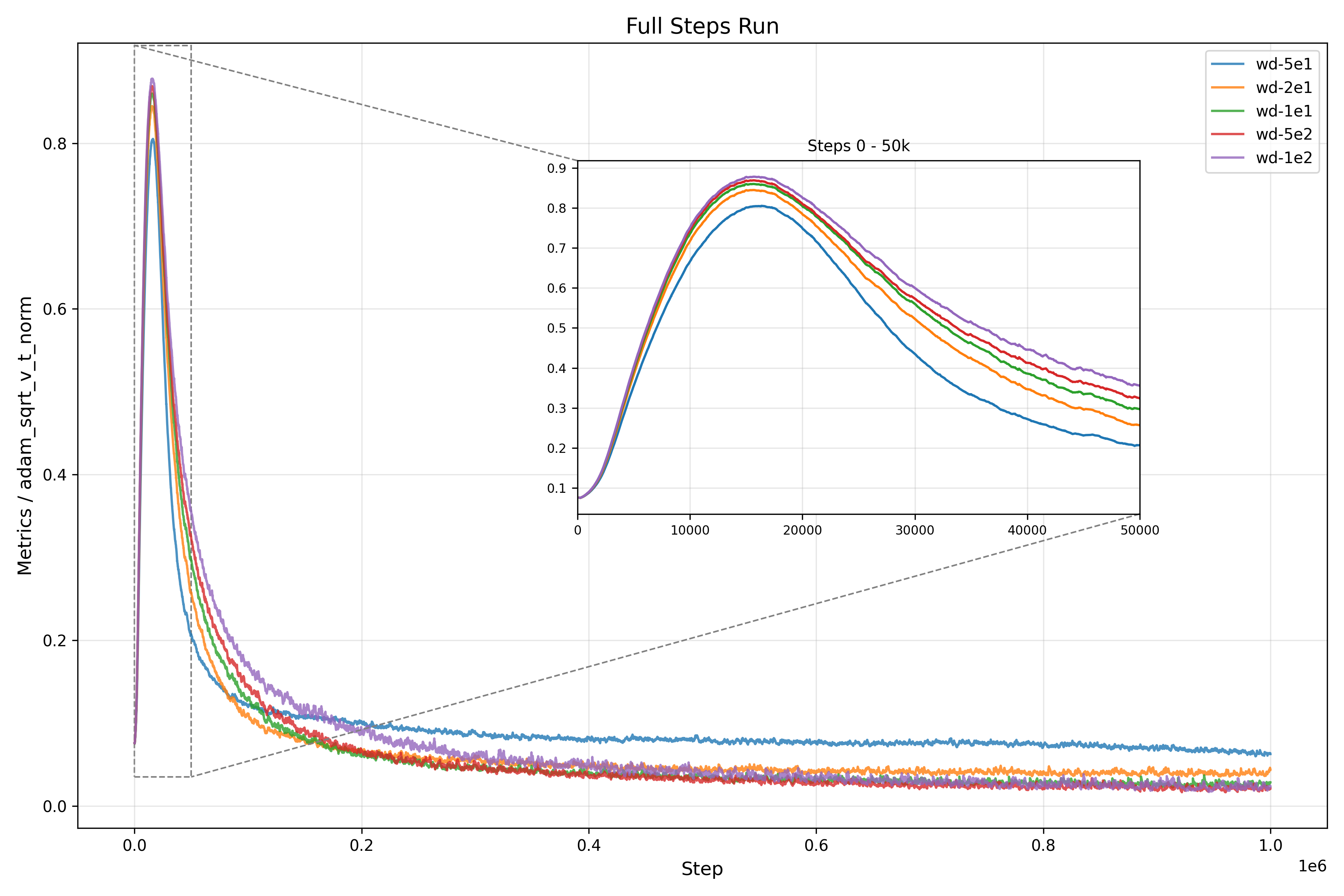}
        \caption{Optimizer Variance $\sqrt{\hat{v}_t}$}
    \end{subfigure}
    \caption{\textbf{Acceleration via Constraint.} (a) Higher weight decay (Peach) significantly accelerates the phase transition. (b) Strong regularization suppresses the peak gradient variance, enforcing a disciplined search in the tangent space.}
    \label{fig:equilibrium_speed}
\end{figure}

Distinct stationary states emerge under analysis (Figure \ref{fig:equilibrium_forces}). High weight decay ($\lambda=0.5$) forces the system into a high-energy, low-radius orbit (Figure \ref{fig:equilibrium_forces}c). This compression accelerates grokking (Figure \ref{fig:equilibrium_speed}) by restricting the search space volume. Effectively, the constraint forces the optimizer to rectify noise into the tangent directions of the manifold, consistent with Riemannian Norm Minimization theories \cite{pesme2025theoretical}.

\subsection{Stability-Regulated Diffusion}
\label{sec:epsilon_ablation}

\begin{figure}[H]
    \centering
    \includegraphics[width=0.8\linewidth]{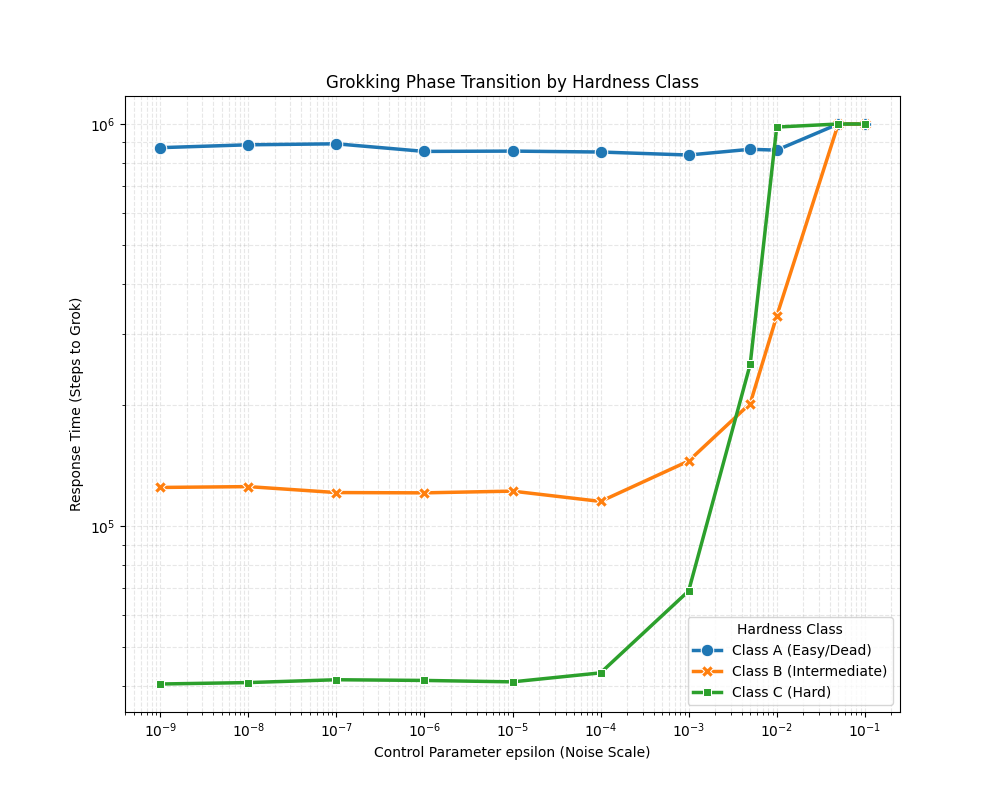} 
    \caption{\textbf{The Stability Threshold.} \textbf{Class C (Hard):} Performance degrades as $\epsilon \to 10^{-1}$, confirming the need for anisotropic noise (Limit I). \textbf{Class B (Intermediate):} Exhibits a convex "optimal stability" region around $\epsilon=10^{-4}$. \textbf{Class A (Easy):} Fails regardless of stability settings.}
    \label{fig:epsilon_grouped}
\end{figure}

The stability parameter $\epsilon$ acts as a spectral gatekeeper. Figure \ref{fig:epsilon_grouped} demonstrates the sensitivity of the phase transition to $\epsilon$.

For solvable tasks, we identify an **Optimal Stability Interval** ($\epsilon \approx 10^{-4}$):
\begin{itemize}
    \item \textbf{Over-Damping ($\epsilon \to 10^{-1}$):} The effective diffusivity $\mathcal{D}_{eff} \to 0$. The optimizer behaves like SGD, failing to explore the manifold.
    \item \textbf{Under-Damping ($\epsilon \to 10^{-9}$):} The system suffers from radial instability. While diffusion is maximized, the lack of a stability floor prevents weight condensation onto the sparse circuit \cite{prieto2025grokking}.
\end{itemize}

\subsection{Mechanism: Spectral Gating and Sharpness Violation}
\label{sec:spectral_gating}

We provide direct evidence that a **Spectral Stability Condition ** governs the delayed generalization. According to Edge of Stability (EoS) theory \cite{cohen2021gradient}, the optimizer cannot stably enter a basin where curvature $\lambda_{max}^H$ exceeds the stability threshold $2/\eta_{eff}$.

\begin{figure}[H]
    \centering
    \includegraphics[width=0.95\linewidth]{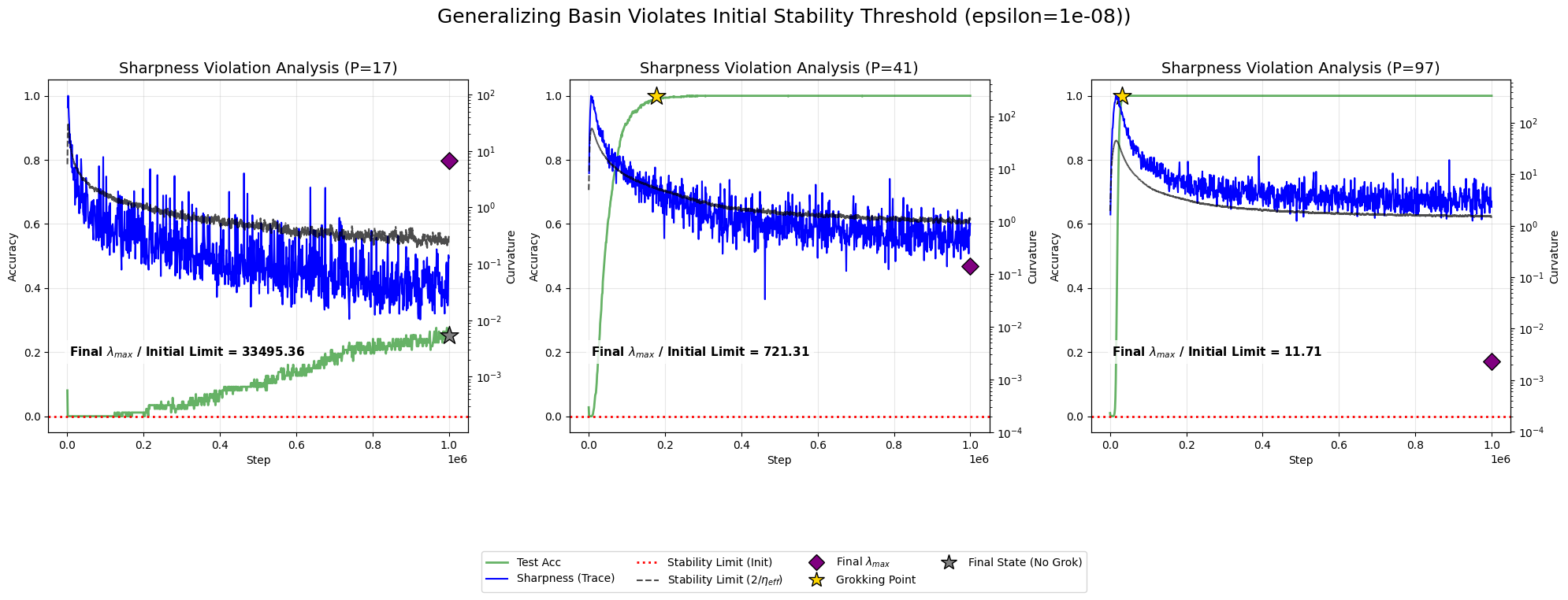} 
    \caption{\textbf{Violation of the Initial Stability Condition.} We compare Hessian Sharpness (Blue) to the \textit{Initial} Stability Limit determined by $\epsilon$ (Red Dotted Line). For the grokking task ($P=41$, Center), the generalizing solution resides in a basin where $\lambda_{max}$ is **721x higher** than the initial stability threshold. The delayed phase (0--200k steps) corresponds to the time required for gradient variance $\sqrt{v_t}$ to grow sufficiently to lift the stability ceiling (Black Dashed Line) above the basin's curvature.}
    \label{fig:sharpness_violation}
\end{figure}

Figure \ref{fig:sharpness_violation} illustrates the core mechanism. For the grokking task ($P=41$), the generalizing minimum is extremely sharp. The "Sharpness Violation Ratio" reaches **721.31x**. This creates a **Spectral Gate**: the model is initially locked out of the generalizing basin. It remains trapped in the flat memorization basin until the accumulated gradient noise variance $\sqrt{v_t}$ increases the denominator of the effective step size. This variance accumulation raises the effective stability ceiling $2/\eta_{eff}$, eventually permitting entry into the sharp "Clock Circuit" basin.

\subsection{The Necessity of Anisotropy: Failure of Isotropic Diffusion}
\label{sec:sgd_failure}

To disentangle the role of "thermal energy" (variance) from "geometric rectification" (covariance), we trained with Stochastic Gradient Langevin Dynamics (SGLD). We injected isotropic Gaussian noise matched to the thermal scale of AdamW.

\begin{figure}[H]
    \centering
    \begin{subfigure}{0.48\textwidth}
        \centering
        \includegraphics[width=\linewidth]{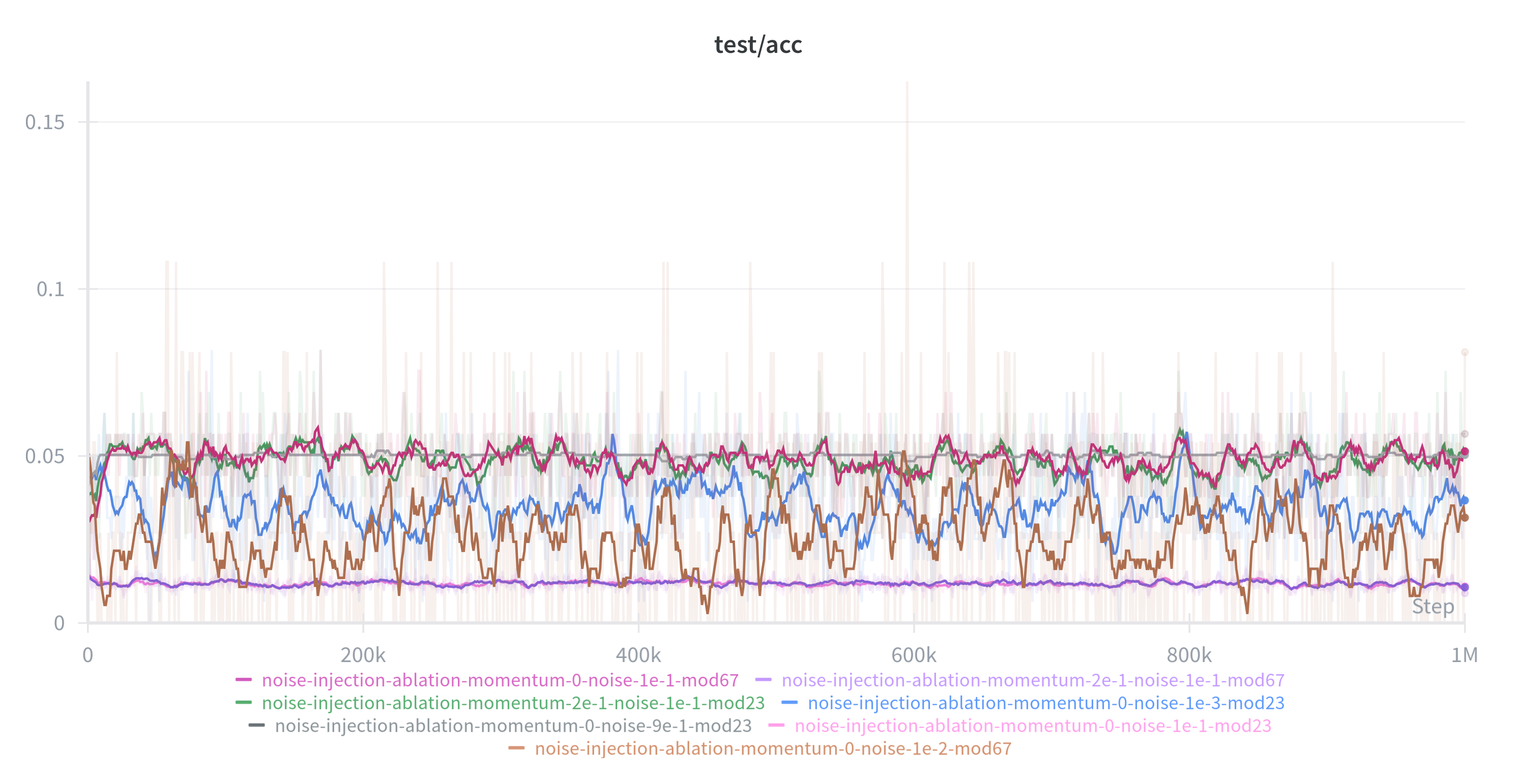} 
        \caption{\texttt{test/acc}: Flatline at Random Chance}
    \end{subfigure}
    \hfill
    \begin{subfigure}{0.48\textwidth}
        \centering
        \includegraphics[width=\linewidth]{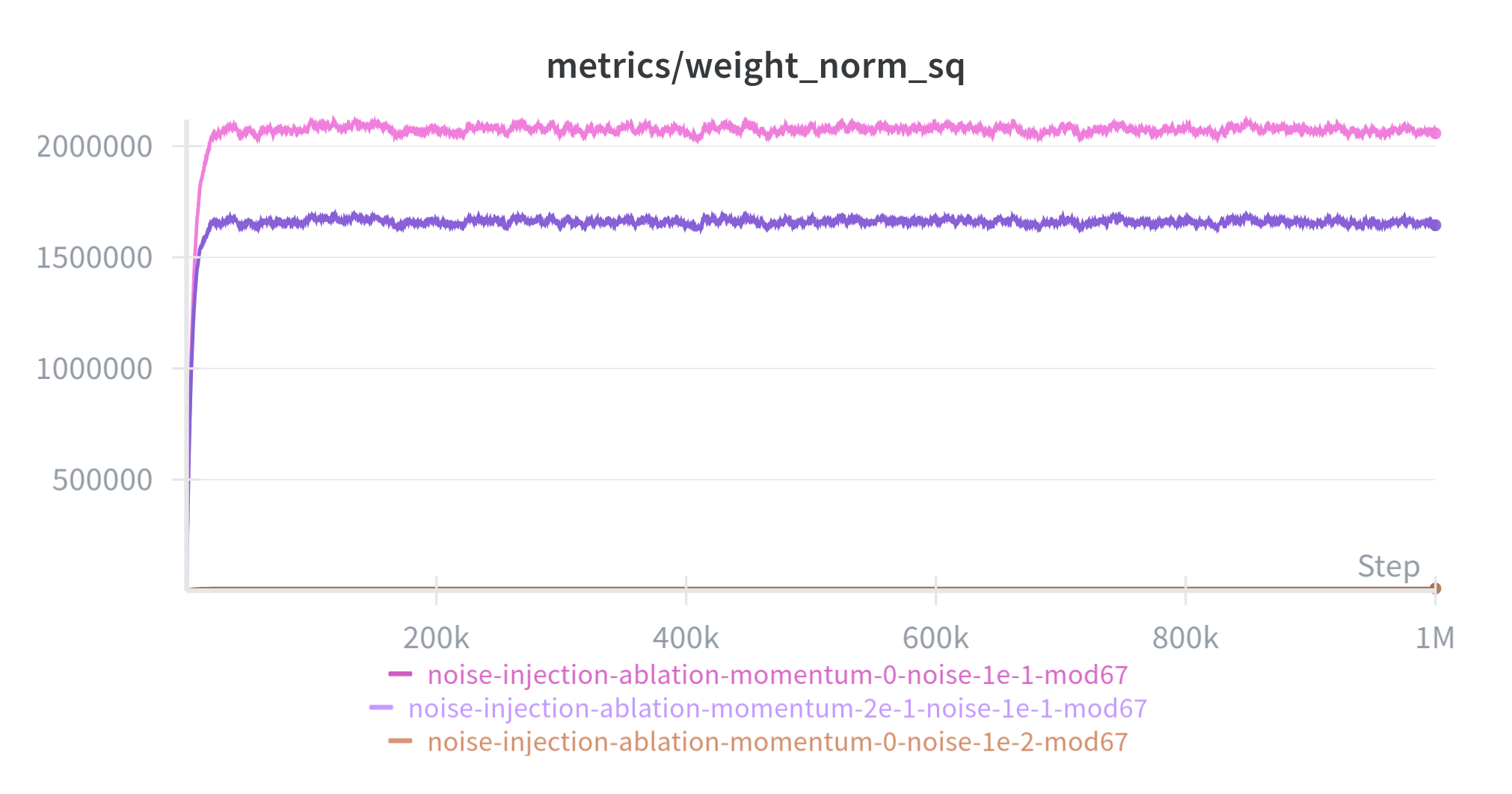} 
        \caption{\texttt{metrics/weight\_norm\_sq}: Scalar Equilibrium}
    \end{subfigure}
    \caption{\textbf{Failure of Isotropic SGLD.} (a) Despite $1 \times 10^6$ steps and extensive noise tuning ($\sigma \in [10^{-3}, 10^{-1}]$), SGLD fails to generalize. (b) Isotropic noise successfully maintains the weight norm, proving that radial energy alone is insufficient.}
    \label{fig:sgd_dynamics}
\end{figure}

\begin{figure}[H]
    \centering
    \begin{subfigure}{0.32\textwidth}
        \centering
        \includegraphics[width=\linewidth]{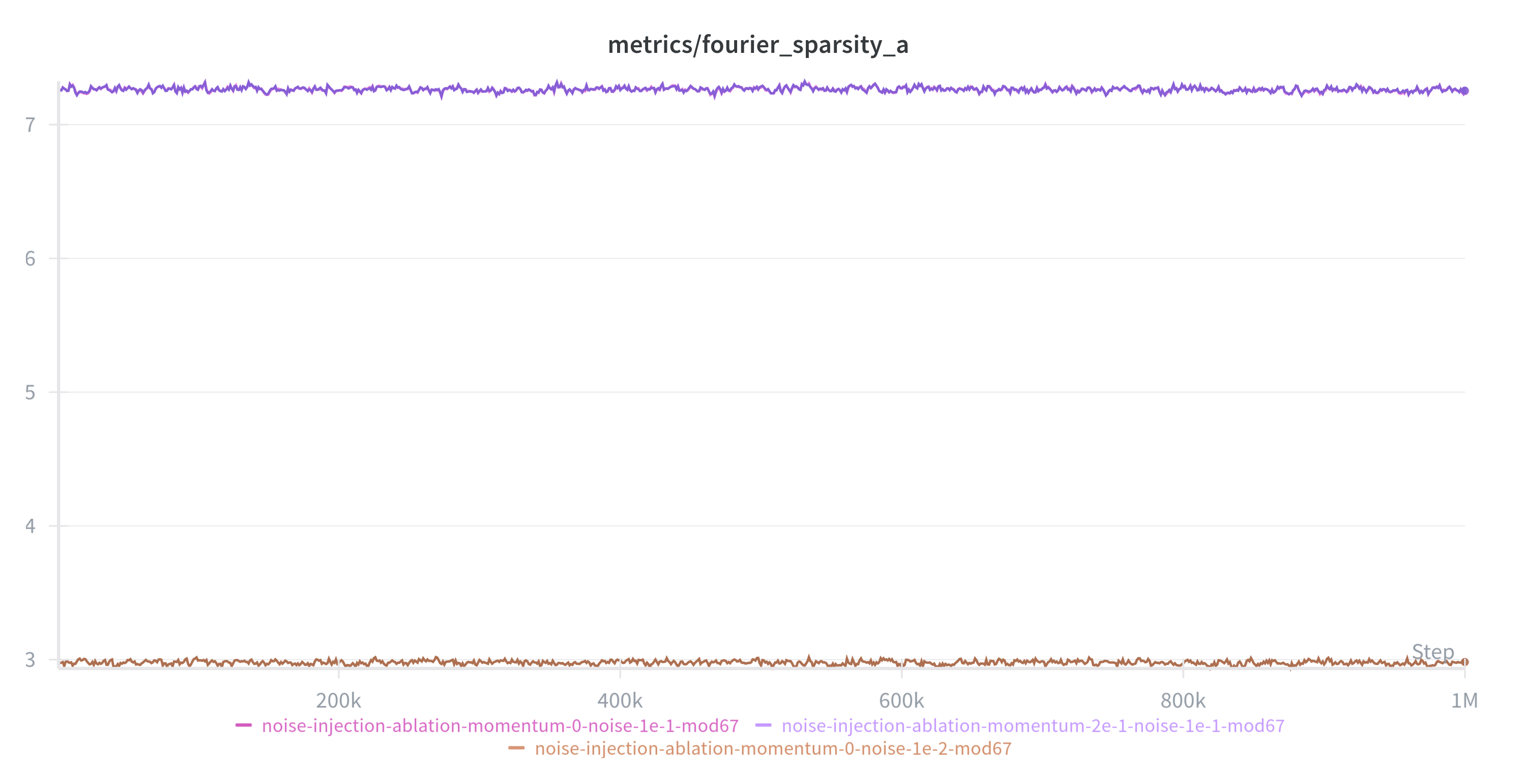} 
        \caption{\texttt{metrics/fourier\_sparsity\_a}: High Entropy}
    \end{subfigure}
    \hfill
        \begin{subfigure}{0.32\textwidth}
        \centering
        \includegraphics[width=\linewidth]{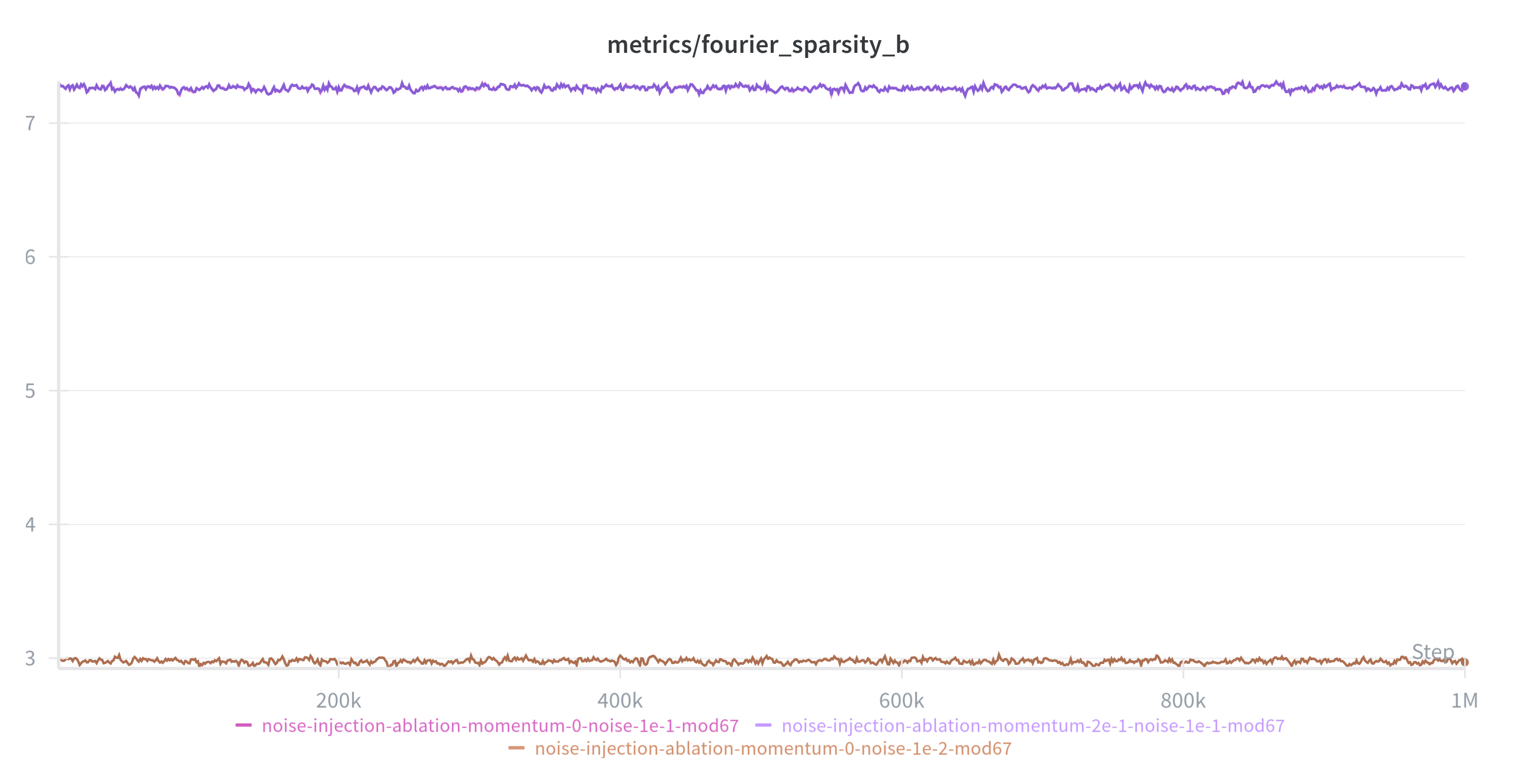} 
        \caption{\texttt{metrics/fourier\_sparsity\_b}: High Entropy}
    \end{subfigure}
    \hfill
    \begin{subfigure}{0.32\textwidth}
        \centering
        \includegraphics[width=\linewidth]{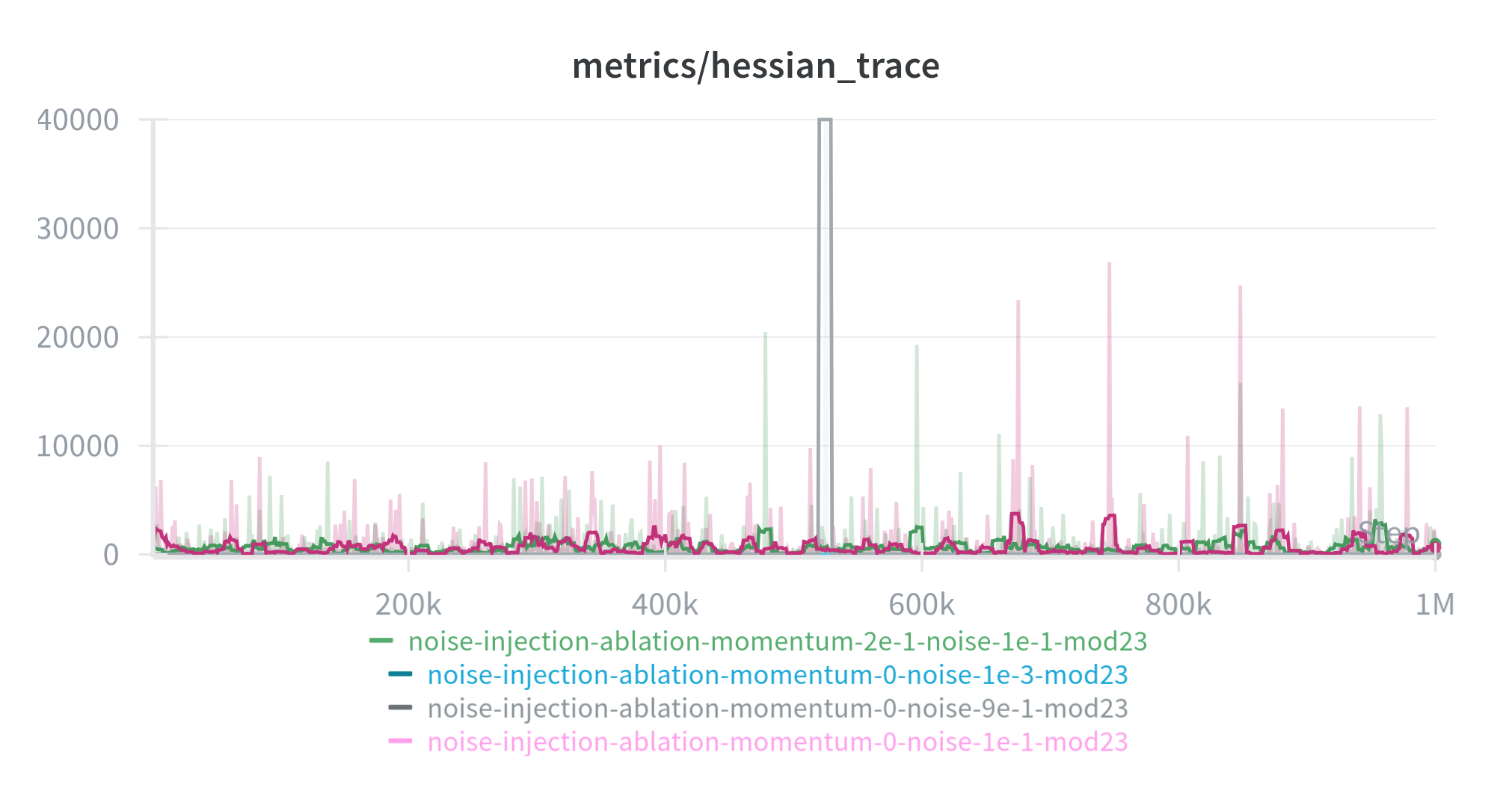} 
        \caption{\texttt{metrics/hessian\_trace}: No Spectral Features}
    \end{subfigure}
    \caption{\textbf{Structural Stagnation in SGLD.} (a) and (b) Fourier sparsity remains high, indicating no circuit formation. (c) The Hessian trace lacks the characteristic "Ignition" spike of AdamW.}
    \label{fig:sgd_structure}
\end{figure}

Isotropic diffusion fails to induce grokking (Figure \ref{fig:sgd_dynamics}). While the noise prevents weight collapse, the model remains in a high-entropy state (Figure \ref{fig:sgd_structure}). This confirms that grokking requires **Anisotropic Rectification**: the optimizer must strictly suppress noise in high-curvature directions while amplifying it in tangent directions to navigate the Minimizing Level Set \cite{balles2018dissecting}.

\section{Discussion}

Grokking has historically been framed as a mystery of timescales—specifically, why generalization lags so far behind training convergence. Our findings suggest this is actually a problem of \textbf{spectral accessibility}. By analyzing AdamW as a \textbf{Variance-Gated Stochastic System}, we show that the delay is not merely a slow drift, but a structural requirement for stability. The optimizer must undergo specific internal state changes to access high-curvature basins. This perspective helps resolve three tensions in the current literature.

\subsection{The Kinetic Mechanism: Geometric Rectification}
Recent geometric frameworks \cite{pesme2025theoretical, musat2025geometry} prove that grokking corresponds to Riemannian Norm Minimization on the Minimizing Level Set $\mathcal{Z}$. However, they attribute the delay primarily to the weak magnitude of weight decay.

Our ablation of Isotropic SGLD (Section \ref{sec:sgd_failure}) refines this view. We find that while isotropic noise can prevent weight collapse and lift test accuracy above random chance, it fails to induce perfect generalization. In simple terms, adding random noise helps the model escape shallow local minima, but it does not provide the steering necessary to find the specific generalizing circuit. Because isotropic noise vectors are statistically orthogonal to the low-dimensional solution manifold $\mathcal{Z}$, standard diffusion results in orthogonal oscillations. AdamW resolves this via **Anisotropic Rectification**: by scaling updates by $v_t^{-1/2}$, it directs noise into the tangent directions of the manifold \cite{balles2018dissecting, xie2021diffusion}. Thus, isotropic noise helps the model move, but geometric rectification ensures it moves in the right direction.

\subsection{Unifying the Stability Paradox}
There is a conflict in the literature regarding instability. Thilak et al. \cite{thilak2022slingshot} argue that "Slingshots"—spikes in loss and gradient variance—drive grokking. Conversely, Prieto et al. \cite{prieto2025grokking} suggest that numerical instability (like Softmax Collapse) prevents learning.

Our **Spectral Gating Theory** (Eq. 5) suggests these views are compatible. We identify the Slingshot as a necessary variance accumulation event.
\begin{itemize}
    \item \textbf{Mechanism of Entry:} A Slingshot causes a spike in the second moment estimator $v_t$. This increase in the denominator reduces the effective step size $\eta_{eff} \approx \eta / \sqrt{v_t}$. This reduction raises the stability ceiling $2/\eta_{eff}$, temporarily allowing the optimizer to survive in the sharper generalizing basin.
    \item \textbf{The Epsilon Constraint:} As noted by Thilak et al., increasing $\epsilon$ halts Slingshots. We derive this analytically: if $\epsilon$ is too high, the variance $v_t$ never dominates the denominator, and the stability ceiling never lifts. If $\epsilon$ is too low, the instability becomes unbounded.
\end{itemize}
Grokking appears to happen near the edge of stability. The optimizer needs enough instability to generate variance for exploration, but too much instability leads to numerical divergence.

\subsection{Revisiting Flatness: The Geometry of Algorithmic Precision}
\label{sec:flatness_revisited}

Kumar et al. \cite{kumar2024grokking} describe grokking as a transition from "Lazy" to "Rich" dynamics. This creates a theoretical tension with the prevailing "Flat Minima" hypothesis \cite{hochreiter1997flat}, which posits that generalization correlates with broad, low-curvature basins.

We propose a topological resolution specific to algorithmic tasks. While flat minima are robust to input noise in perceptual tasks (e.g., image classification), algorithmic solutions—such as the trigonometric "Clock Circuit" identified by Nanda et al. \cite{nanda2023progress}—rely on **precise interference patterns**. Constructive interference is geometrically fragile: small perturbations in the frequency or phase parameters destroy the circuit's logic. Consequently, the generalizing solution resides in a **narrow, high-curvature manifold**.

In contrast, the "Lazy" memorization solution relies on a look-up table structure. This basin is **entropically vast**: there are combinatorially many ways to assign weights to memorized disconnected points, creating a broad, flat attractor that is easily accessible from random initialization.

The grokking delay is therefore the kinetic cost of this geometry. The optimizer is initially trapped in the high-entropy memorization basin. It cannot access the low-entropy, high-precision generalizing manifold until it accumulates sufficient gradient variance to lift the spectral gate (Eq. 5), satisfying the strict stability requirements of the algorithmic solution.

\subsection{Dimensional Bottlenecks: Capacity Collapse}
\label{sec:capacity_collapse}

\begin{figure}[H]
    \centering
    \includegraphics[width=0.8\linewidth]{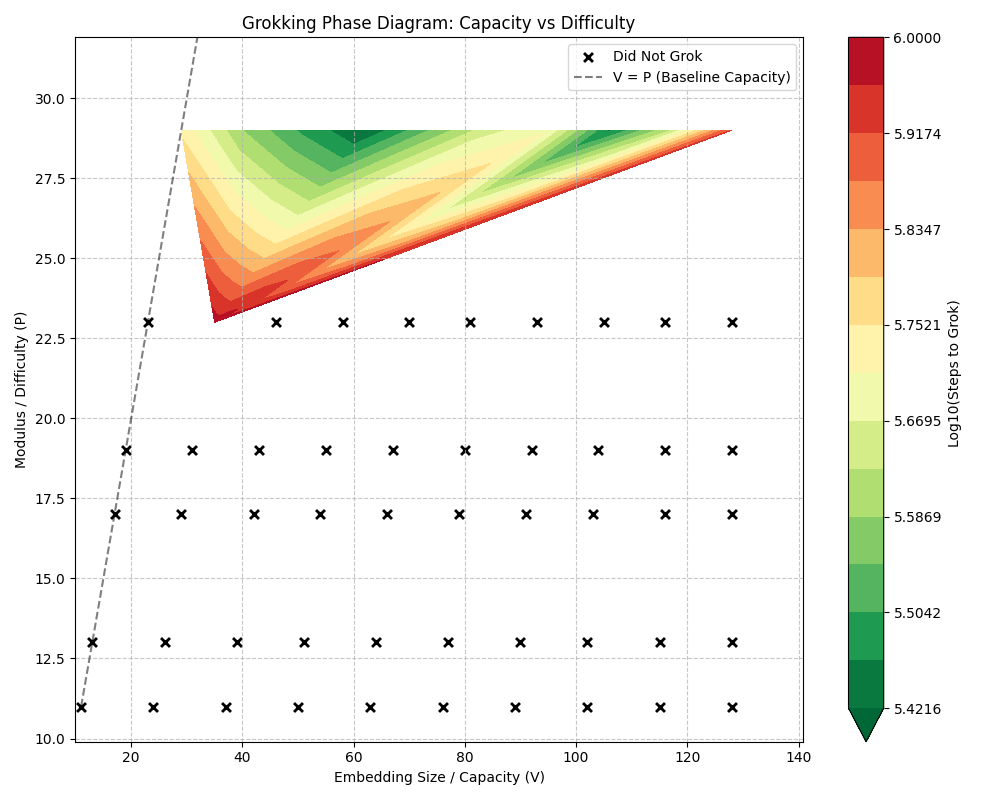}
    \caption{\textbf{Capacity Collapse.} We ablate embedding dimension $V$ against modulus $P$. The color scale indicates the time required to grok (Log10 Steps). The sharp diagonal boundary reveals that grokking requires a minimum dimensional capacity proportional to task complexity ($V \propto P$). Black crosses indicate configurations that failed to generalize within $10^6$ steps.}
    \label{fig:capacity_collapse}
\end{figure}

Finally, we address the limits of this phenomenon. Our embedding ablation (Figure \ref{fig:capacity_collapse}) reveals that task difficulty is governed not just by the modulus $P$, but by the ratio of capacity to complexity.

As shown in Figure \ref{fig:capacity_collapse}, a sharp phase boundary emerges. Failure occurs when the embedding dimension is insufficient relative to the modulus ($V \lesssim P$). We term this **Capacity Collapse**. When the embedding space lacks the geometric bandwidth to represent the $P$ discrete Fourier modes required for the solution \cite{liu2022omnigrok}, the gradient signal becomes **rank-deficient**. This prevents the optimizer from constructing the generalizing circuit regardless of the variance supplied. This finding confirms that spectral gating is a secondary mechanism that can only operate when the model capacity is geometrically sufficient to support the solution.

\section{Conclusion}

We have characterized grokking as a conditional phase transition governed by the interplay between landscape geometry and optimizer stability. We establish three governing principles:

\begin{enumerate}
    \item \textbf{The Complexity Threshold:} Generalization is not monotonic. We identify a "Signal Starvation" regime where the landscape is structurally barren ($V < P$), and a "Complexity Override" regime ($P > 67$) where memorization becomes dimensionally unstable, forcing early generalization.
    
    \item \textbf{Spectral Gating:} We demonstrate that the delayed phase is a stable equilibrium where the model is spectrally barred from the generalizing basin. The delay represents the time required for gradient variance to accumulate and lift the stability ceiling defined by $\epsilon$.
    
    \item \textbf{Anisotropic Rectification:} We show that thermal energy alone is insufficient. Generalization requires the directed channeling of variance unique to adaptive optimizers, which rectify noise into the tangent space of the solution manifold.
\end{enumerate}

In short, generalization depends on how the optimizer balances noise and stability. Future work could investigate whether similar spectral gating mechanisms govern the emergence of reasoning capabilities in larger language models, or if this behavior is specific to highly structured algorithmic tasks.

\appendix

\section{Appendix}

This supplementary material provides detailed geometric and spectral evidence supporting the **Spectral Gating** framework presented in the main text. We provide topological visualizations of the embedding space (Section \ref{app:topology}), quantify the anisotropic covariance structure of AdamW updates (Section \ref{app:anisotropy}), and examine the spectral limits of thermal noise injection across the full range of task complexities (Section \ref{app:thermal_assist}).

\subsection{Topological Analysis: The Clock Face}
\label{app:topology}

\begin{figure}[H]
    \centering
    \includegraphics[width=\linewidth]{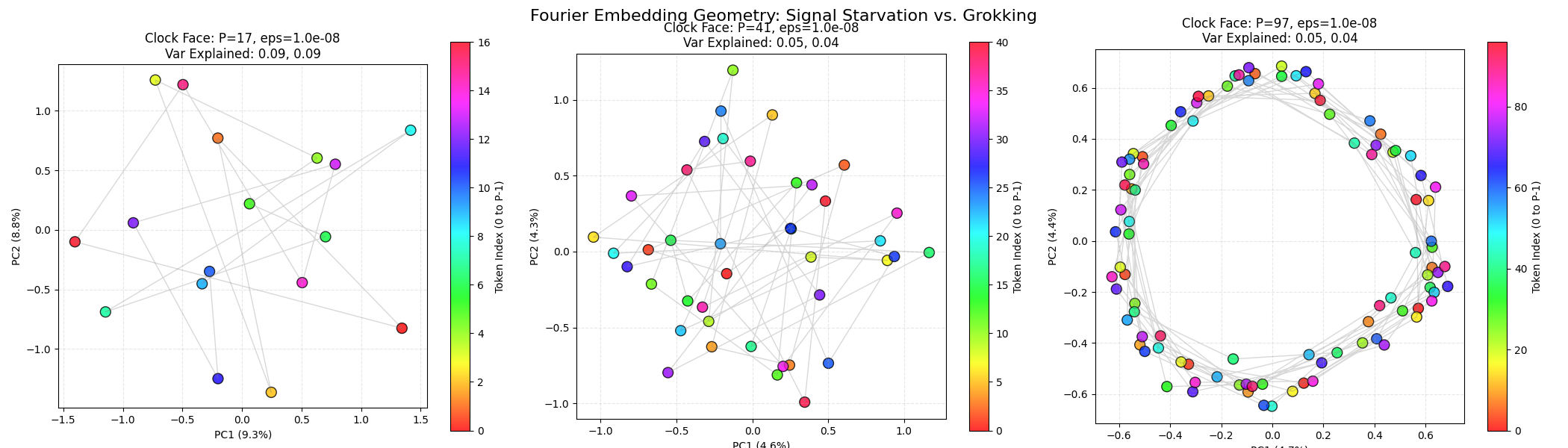}
    \caption{\textbf{Embedding Geometry across Complexity Regimes.} Learned embeddings projected onto their first two principal components. 
    \textbf{Left ($P=17$):} The Signal Starvation regime exhibits a lack of structural organization; embeddings are scattered, indicating failure to discover the modular circle.
    \textbf{Center ($P=41$):} The Grokking regime reveals a recognizable but noisy circular geometry, consistent with a variance-limited equilibrium where the optimizer maintains high entropy to navigate the basin.
    \textbf{Right ($P=97$):} The Stability Override regime forms a clear circular structure immediately, indicating that high-complexity tasks force rapid geometric condensation.}
    \label{fig:clock_faces}
\end{figure}

Section \ref{sec:phase_diagram} hypothesized that the "Signal Starvation" regime ($P < 23$) stems from a failure to form the requisite task geometry. Figure \ref{fig:clock_faces} visualizes the embedding weights $W_E$ via Principal Component Analysis (PCA).

\subsection{Evidence of Anisotropy}
\label{app:anisotropy}

\begin{figure}[H]
    \centering
    \begin{subfigure}{0.48\textwidth}
        \centering
        \includegraphics[width=\linewidth]{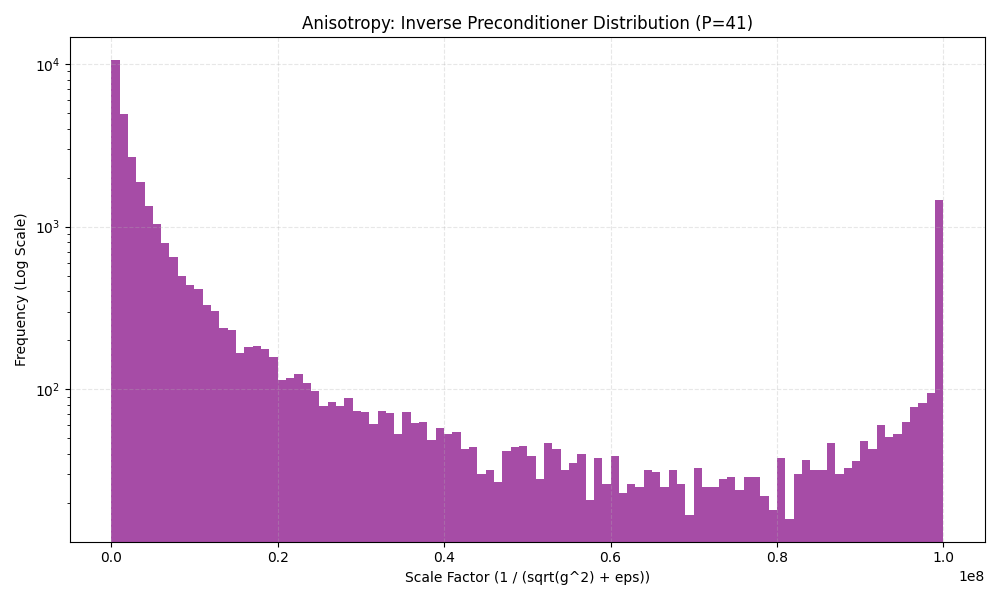}
        \caption{Grokking Regime ($P=41$)}
    \end{subfigure}
    \hfill
    \begin{subfigure}{0.48\textwidth}
        \centering
        \includegraphics[width=\linewidth]{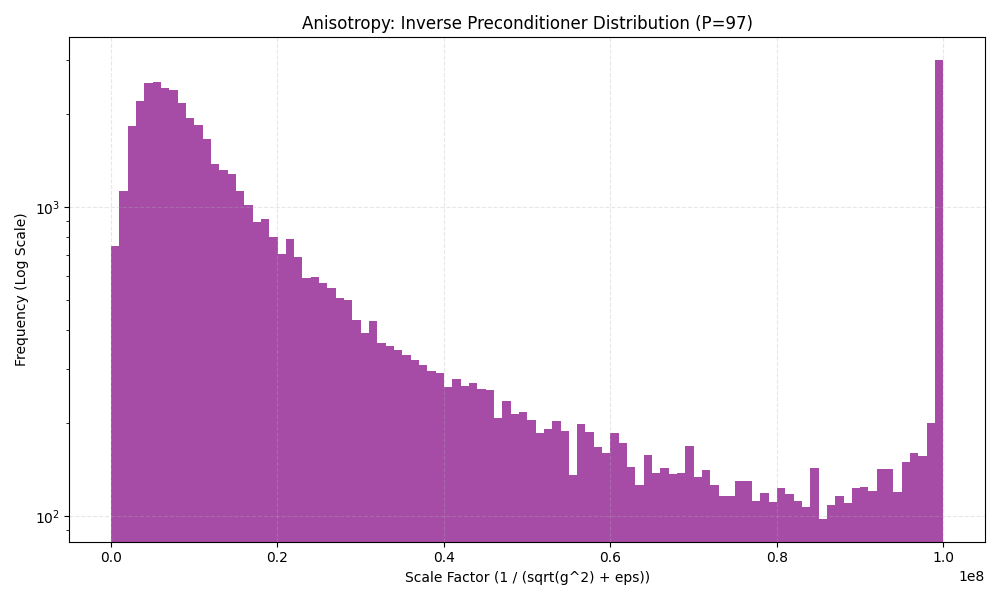}
        \caption{Override Regime ($P=97$)}
    \end{subfigure}
    \caption{\textbf{Distribution of Update Scalings.} Histograms display the effective learning rates applied by AdamW per parameter. Unlike Isotropic SGLD (which implies a Dirac distribution), AdamW generates a **heavy-tailed distribution** spanning multiple orders of magnitude. This indicates that the optimizer actively shapes the noise geometry, suppressing variance in high-curvature directions (left tail) while amplifying it in flat directions (right tail).}
    \label{fig:anisotropy}
\end{figure}

Section \ref{sec:sgd_failure} argued that grokking relies on **Anisotropic Rectification**—the selective amplification of noise in tangent directions. Figure \ref{fig:anisotropy} characterizes this by plotting the distribution of the inverse preconditioner scale factors $\alpha_i = 1/(\sqrt{v_{t,i}} + \epsilon)$ at the moment of generalization.

\subsection{Spectral Dynamics across Complexity}
\label{app:spectral_grid}

\begin{figure}[H]
    \centering
    \includegraphics[width=\linewidth]{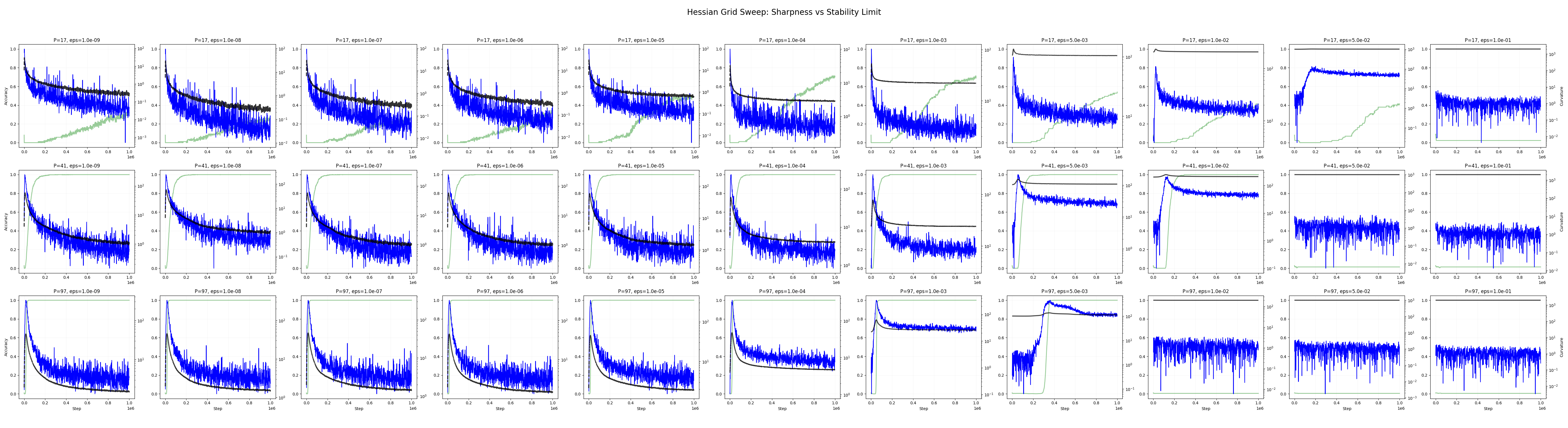}
    \caption{\textbf{Spectral Dynamics for Varying Moduli.} Green lines denote Test Accuracy; Blue lines denote Hessian Sharpness. For successful generalization cases ($P \ge 29$), we observe the characteristic "Ignition" event where the variance-driven stability ceiling (Black Dashed) rises to meet the local curvature.}
    \label{fig:hessian_grid}
\end{figure}

To ensure the mechanisms identified in Section \ref{sec:spectral_gating} apply beyond specific cases, Figure \ref{fig:hessian_grid} presents the spectral dynamics across the full range of moduli.

\subsection{Limits of Isotropic Noise}
\label{app:thermal_assist}

\begin{figure}[H]
    \centering
    \includegraphics[width=0.95\linewidth]{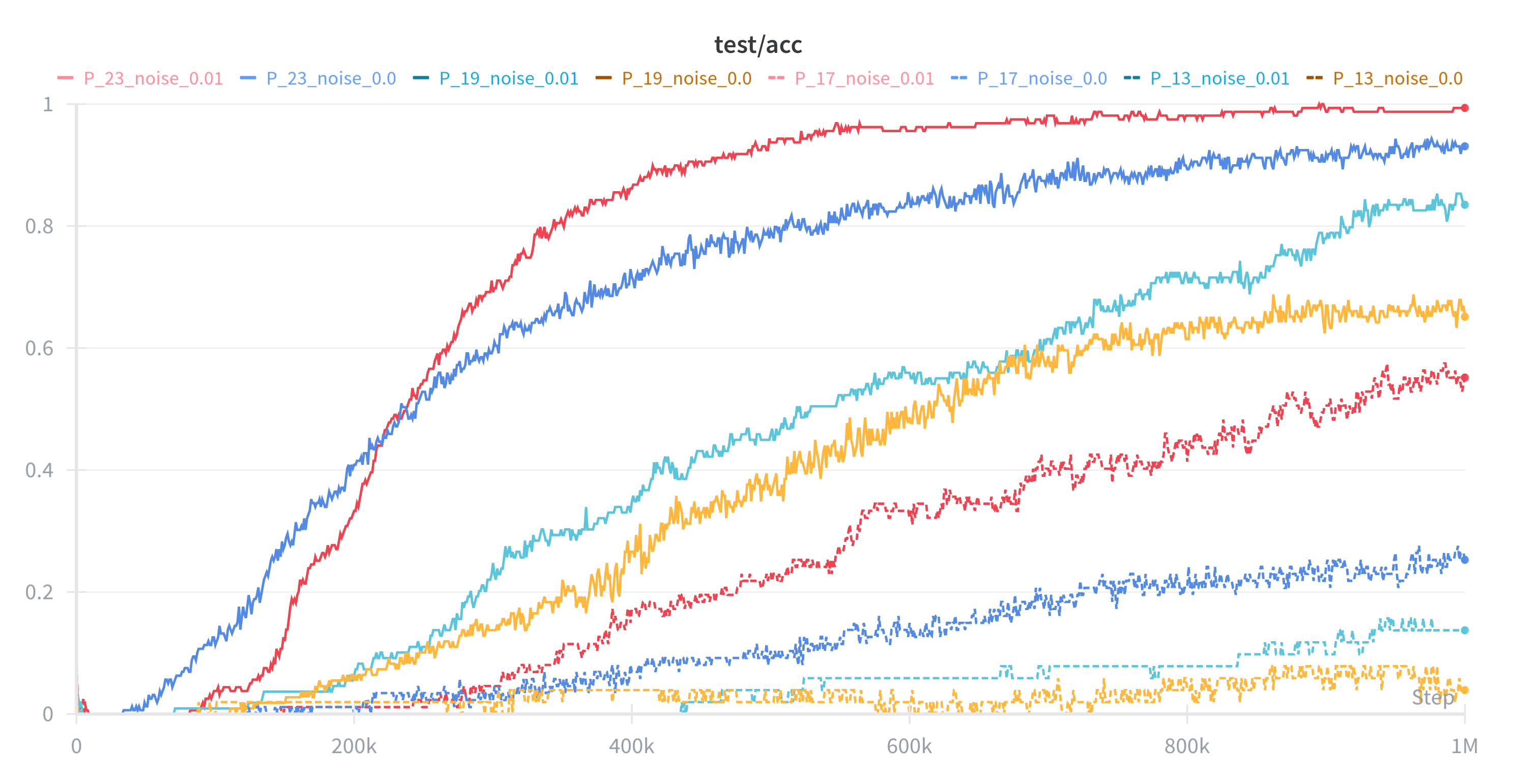}
    \caption{\textbf{Isotropic Noise Injection.} Injecting isotropic noise ($\sigma=0.01$, cyan lines) allows the model to escape the lazy basin and reach $\approx 60-80\%$ accuracy, outperforming the noiseless baseline. However, it fails to induce the perfect generalization seen in grokking. Thus, isotropic noise aids optimization but does not replace the geometric steering required for structural convergence.}
    \label{fig:noise_lines}
\end{figure}

Figure \ref{fig:noise_lines} provides a granular view of the interaction between isotropic noise and the "Signal Starvation" regime.

\begin{thebibliography}{10}

\bibitem{power2022grokking}
Alethea Power, Yuri Burda, Harri Edwards, Igor Babuschkin, and Vedant Misra.
\newblock Grokking: Generalization beyond overfitting on small algorithmic datasets.
\newblock {\em arXiv preprint arXiv:2201.02177}, 2022.

\bibitem{liu2022omnigrok}
Ziming Liu, Eric J. Michaud, and Max Tegmark.
\newblock Omnigrok: Grokking beyond algorithmic data.
\newblock {\em ICLR}, 2023.

\bibitem{nanda2023progress}
Neel Nanda, Lawrence Chan, Tom Liberum, Jess Smith, and Jacob Steinhardt.
\newblock Progress measures for grokking via mechanistic interpretability.
\newblock {\em International Conference on Learning Representations (ICLR)}, 2023.

\bibitem{loshchilov2017decoupled}
Ilya Loshchilov and Frank Hutter.
\newblock Decoupled weight decay regularization.
\newblock {\em ICLR}, 2019.

\bibitem{thilak2022slingshot}
Vimal Thilak, Etai Littwin, Shuangfei Zhai, Omid Saremi, Roni Paiss, and Joshua Susskind.
\newblock The slingshot mechanism: An empirical study of adaptive optimizers and the grokking phenomenon.
\newblock {\em arXiv preprint arXiv:2206.04817}, 2022.

\bibitem{cohen2021gradient}
Jeremy Cohen, Simran Kaur, Yuanzhi Li, J. Zico Kolter, and Ameet Talwalkar.
\newblock Gradient descent on neural networks typically occurs at the edge of stability.
\newblock {\em ICLR}, 2021.

\bibitem{musat2025geometry}
Tiberiu Musat.
\newblock The Geometry of Grokking: Norm Minimization on the Zero-Loss Manifold.
\newblock {\em arXiv preprint arXiv:2511.01938}, 2025.

\bibitem{soudry2018implicit}
Daniel Soudry, Elad Hoffer, Mor Shpigel Nacson, Suriya Gunasekar, and Nathan Srebro.
\newblock The implicit bias of gradient descent on separable data.
\newblock {\em The Journal of Machine Learning Research}, 19(1):2822--2878, 2018.

\bibitem{hochreiter1997flat}
Sepp Hochreiter, Jurgen Schmidhuber
\newblock Flat Minima 
\newblock {\em Neural Computation}, 1997

\bibitem{balles2018dissecting}
Lukas Balles and Philipp Hennig.
\newblock Dissecting Adam: The sign, magnitude and variance of stochastic gradients.
\newblock {\em Proceedings of the 35th International Conference on Machine Learning (ICML)}, 2018.

\bibitem{kunstner2019limitations}
Frederik Kunstner, Philipp Hennig, and Lukas Balles.
\newblock Limitations of the empirical fisher approximation for natural gradient descent.
\newblock {\em Advances in Neural Information Processing Systems (NeurIPS)}, 32, 2019.

\bibitem{jelassi2022towards}
Samy Jelassi and Yuanzhi Li.
\newblock Towards understanding how momentum improves generalization in deep learning.
\newblock {\em Proceedings of the 39th International Conference on Machine Learning (ICML)}, 2022.

\bibitem{arora2022scale}
Sanjeev Arora, Zhiyuan Li, and Abhishek Panigrahi.
\newblock The scale of initialization determines the edge of stability.
\newblock {\em Advances in Neural Information Processing Systems (NeurIPS)}, 35, 2022.

\bibitem{damian2023self}
Alex Damian, Eshaan Nichani, and Jason D. Lee.
\newblock Self-stabilization: The implicit bias of gradient descent at the edge of stability.
\newblock {\em International Conference on Learning Representations (ICLR)}, 2023.

\bibitem{simsekli2019tail}
Umut Simsekli, Levent Sagun, and Mert Gurbuzbalaban.
\newblock A tail-index analysis of stochastic gradient noise in deep neural networks.
\newblock {\em Proceedings of the 36th International Conference on Machine Learning (ICML)}, 2019.

\bibitem{xie2021diffusion}
Zeke Xie, Issei Sato, and Masashi Sugiyama.
\newblock A diffusion theory for deep learning dynamics: Stochastic gradient descent exponentially favors flat minima.
\newblock {\em International Conference on Learning Representations (ICLR)}, 2021.

\bibitem{merrill2023tale}
William Merrill, Nikolaos Tsilivis, and Aman Shukla.
\newblock A tale of two circuits: Grokking as competition of sparse and dense subnetworks.
\newblock {\em arXiv preprint arXiv:2303.11873}, 2023.

\bibitem{kumar2024grokking}
Tanishq Kumar, Blake Bordelon, Samuel J. Gershman, and Cengiz Pehlevan.
\newblock Grokking as the transition from lazy to rich training dynamics.
\newblock {\em International Conference on Learning Representations (ICLR)}, 2024.

\bibitem{pesme2025theoretical}
Scott Pesme, Etienne Boursier, and Radu-Alexandru Dragomir.
\newblock A Theoretical Framework for Grokking: Interpolation followed by Riemannian Norm Minimisation.
\newblock {\em Advances in Neural Information Processing Systems (NeurIPS)}, 2025.

\bibitem{varma2023explaining}
Vikrant Varma, Rohin Shah, Zachary Kenton, J{\'a}nos Kram{\'a}r, and Ramana Kumar.
\newblock Explaining grokking through circuit efficiency.
\newblock {\em arXiv preprint arXiv:2309.02390}, 2023.

\bibitem{prieto2025grokking}
Lucas Prieto, Melih Barsbey, Pedro A.M. Mediano, and Tolga Birdal.
\newblock Grokking at the Edge of Numerical Stability.
\newblock {\em International Conference on Learning Representations (ICLR)}, 2025.

\end{thebibliography}
\end{document}